\documentclass{article}

\usepackage{microtype}
\usepackage{graphicx}
\usepackage{subfigure}
\usepackage{booktabs} 

\usepackage{hyperref}

\usepackage[accepted]{synsml2023}

\usepackage{amsmath}
\usepackage{amssymb}
\usepackage{mathtools}
\usepackage{amsthm}

\usepackage[capitalize,noabbrev]{cleveref}

\theoremstyle{plain}
\newtheorem{theorem}{Theorem}[section]
\newtheorem{proposition}[theorem]{Proposition}
\newtheorem{lemma}[theorem]{Lemma}
\newtheorem{corollary}[theorem]{Corollary}
\theoremstyle{definition}
\newtheorem{definition}[theorem]{Definition}
\newtheorem{assumption}[theorem]{Assumption}
\theoremstyle{remark}
\newtheorem{remark}[theorem]{Remark}

\usepackage[textsize=tiny]{todonotes}

\synsmltitlerunning{Submission and Formatting Instructions for Syns \& ML at ICML 2023}

\begin{document}

\twocolumn[
\synsmltitle{Neural Polytopes}





\begin{synsmlauthorlist}
\synsmlauthor{Koji Hashimoto}{yyy}
\synsmlauthor{Tomoya Naito}{yyy3,yyy4}
\synsmlauthor{Hisashi Naito}{yyy2}
\end{synsmlauthorlist}

\synsmlaffiliation{yyy}{Department of Physics, Kyoto University, Kyoto, Japan}
\synsmlaffiliation{yyy3}{RIKEN Interdisciplinary Theoretical and Mathematical Sciences Program (iTHEMS), Wako, Japan}
\synsmlaffiliation{yyy4}{Department of Physics, The University of Tokyo, Tokyo, Japan}
\synsmlaffiliation{yyy2}{Graduate School of Mathematics, Nagoya University, Nagoya, Japan}

\synsmlcorrespondingauthor{Koji Hashimoto}{koji@scphys.kyoto-u.ac.jp}

\synsmlkeywords{Polytopes, polyhedra, polygons}

\vskip 0.3in
]



\printAffiliationsAndNotice{}  

\begin{abstract}
We find that simple neural networks with ReLU activation generate polytopes as an approximation of a unit sphere in various dimensions. 
The species of polytopes are regulated by the network architecture, such as the number of units and layers. For a variety of activation functions, generalization of polytopes is obtained, which we call neural polytopes. They are a smooth analogue of polytopes, exhibiting geometric duality. 
This finding initiates research of generative discrete geometry to approximate surfaces by machine learning. 

\end{abstract}

\section*{Introduction}

The approach humans have taken to modeling nature is to approximate smooth curved surfaces in nature with linear objects such as planes. This is natural in terms of recognizing natural objects composed of smooth curves, since the simplest solution to the equations of motion in physics is linear motion with constant velocity. In ancient Greece, polygons were discovered as a way to approximate a circle by straight line segments, and polyhedra as a way to approximate a sphere by pieces of planes, and in particular, it was assumed by Plato that the existence of only a finite number of regular polyhedra was the fundamental understanding of nature. Thus, approximating rotationally symmetric objects (circles and spheres) with piecewise linear functions is at the root of modeling nature.

Machine learning, on the other hand, is known to be able to approximate any function if the network architecture is multi-unit multilayer, as stated in the universal approximation theorem \cite{cybenko1989approximation, hornik1989multilayer, hornik1991approximation}. In particular, taking the activation function to be a step function or ReLU can be identified with approximation by a piecewise constant or piecewise linear function. Therefore, the modeling of approximation of natural phenomena by neural networks should naturally lead us to the rediscovery of discrete geometry, when we go back to the motivation in ancient Greek times and target the rotationally symmetric objects. 

Discrete geometry is the research field in mathematics to develop methods for discretizing smooth surfaces, whose application ranges from computer graphics \cite{baumgart1974geometric} to quantum physics. To cite an example of the latter, the fundamental goal of completing the quantum theory of gravity in physics is to discretely approximate and quantize a smoothly curved surface called spacetime \cite{Regge:1961px}. Thus, the development of methods in discrete geometry is long overdue.

In this study, a sphere is approximated by a neural network function, as a first numerical experiment to bridge discrete geometry and machine learning and to explore possible visualization of trained functions.

As we describe below, for the choice of the activation function as ReLU, polygons and polytopes are naturally
generated. When we allow other activation functions, we obtain infinite families of generalization of polytopes --- which we call \textit{neural polytopes}.

\section*{Method}

A plane in the Euclidean $d$-dimensional space spanned by the coordinate $(x_1, \ldots, x_d)$ is parameterized as
\begin{align}
    \sum_i a_i x_i = 1,
    \label{linfun}
\end{align}
where $a_i$ ($i=1$, \ldots, $d$) are real constant parameters. Polyhedra are nothing but a generalization of this equation to a piecewise linear function. Note that the right-hand side of \eqref{linfun} needs to be fixed to be the unity; otherwise, we have to require some affine quotient.

\begin{figure}[t]
\begin{center}
\includegraphics[width=1\columnwidth]{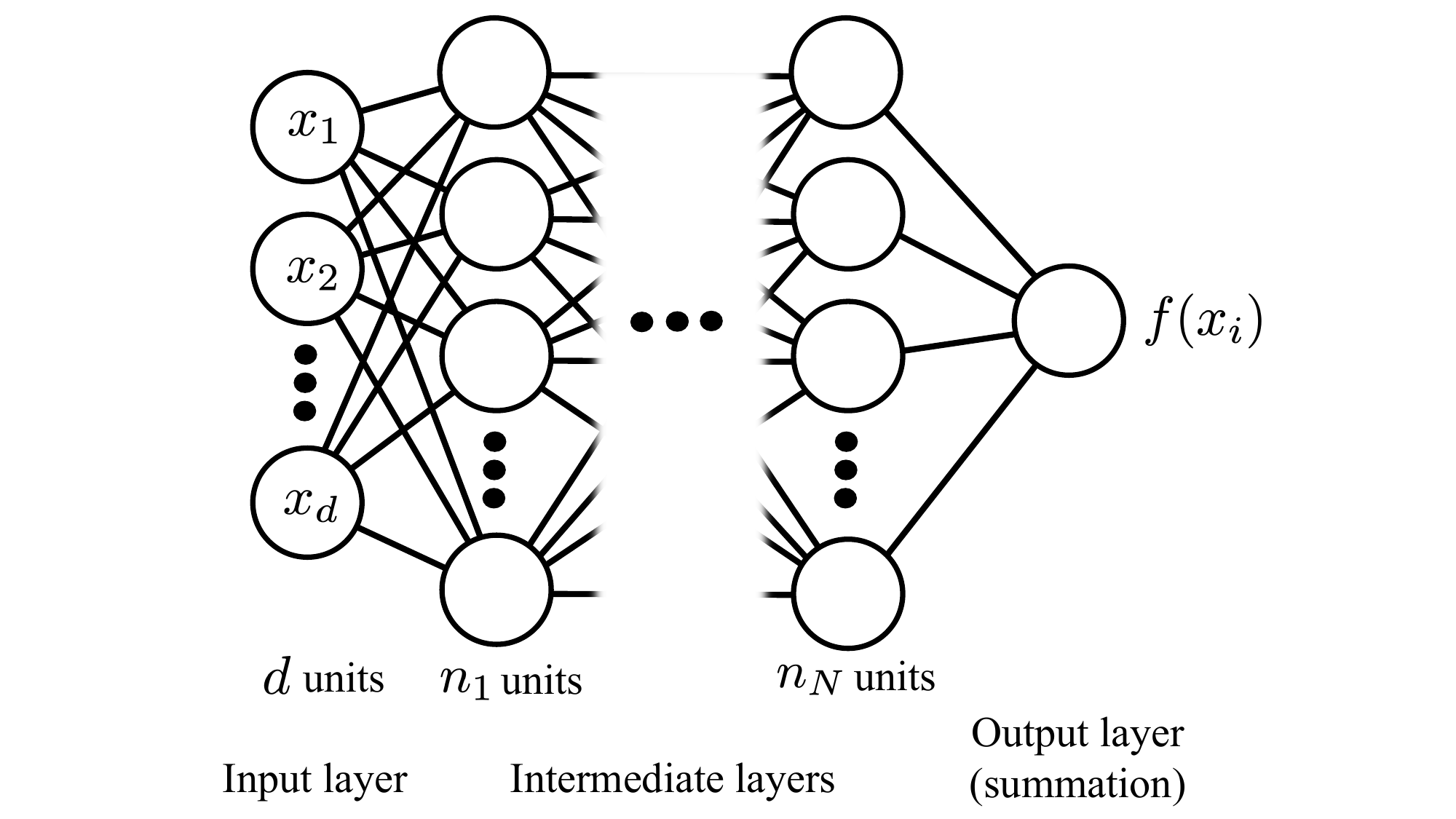}
\vspace*{-5mm}
\caption{The architecture of the neural network used in our work.}
\label{fig0}
\end{center}
\end{figure}

Now, we notice that any feed-forward deep neural network with the ReLU activation function without the bias is exactly of the form of the left-hand side of \eqref{linfun}.
Therefore, let us prepare a deep neural network architecture with $N$ intermediate fully-connected layers with $(n_1, n_2, \ldots, n_N)$ units each, and with the ReLU activation function. The input layer consists of $d$ units whose input is just the coordinate $\vec{x}=(x_1, \ldots, x_d)$. For simplicity, the output layer is taken to be a summation layer which sums the values of the $n_N$ units at the last intermediate layer. We force all biases equal to be zero. See Fig.~\ref{fig0} for the architecture.

The preparation of the training data is straightforward; since we aim to approximate the $(d-1)$-sphere, we produce a set of random points on the $(d-1)$-sphere in the Cartesian coordinates, and produce the data set ${\cal D}$ of the form
\begin{align}
    {\cal D} \equiv \{\vec{x}^{(i)}\to 1 \mid \vec{x}^{(i)}\in S^{d-1} \}.
\end{align}
That is, the input is a random point on the $S^{d-1}$, and the output is the unity.

The activation function $\varphi(x)$ 
other than ReLU,
\begin{align}
    \varphi(x) = \left| x \right|^p
    \label{symact}
\end{align}
with a positive real constant $p$, gives geometrically symmetric neural network functions, as it respects the reflection symmetry $x \to -x$. This paper focuses on the results with \eqref{symact} for a better symmetric approximation of spheres. 
Note that choosing $p=2$ results in the complete reproduction of the original sphere (circle), since the equation to define the sphere is of the symmetric quadratic functions, $\sum_i (x_i)^2 = 1$.

\begin{figure}[t]
\begin{center}
\includegraphics[width=0.3\columnwidth]{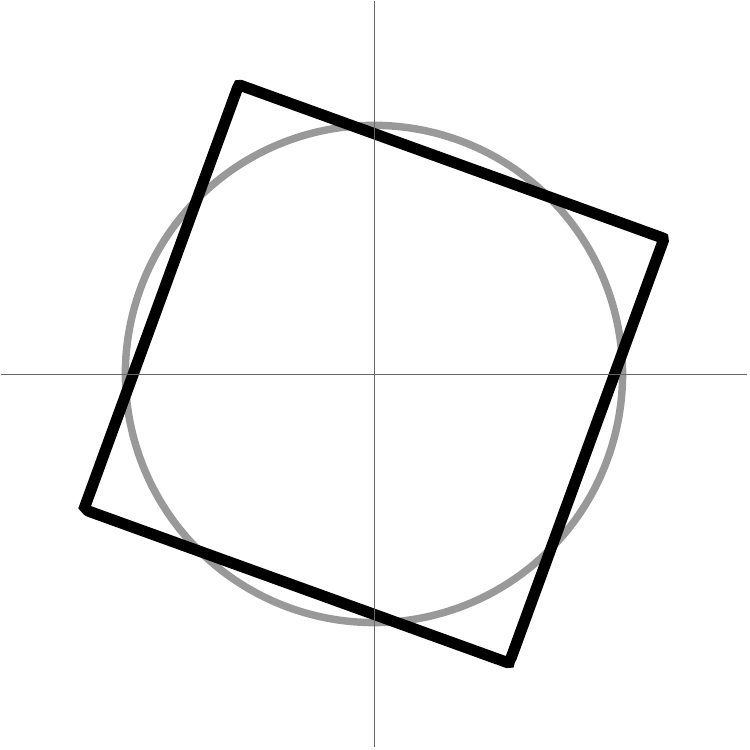}
\includegraphics[width=0.3\columnwidth]{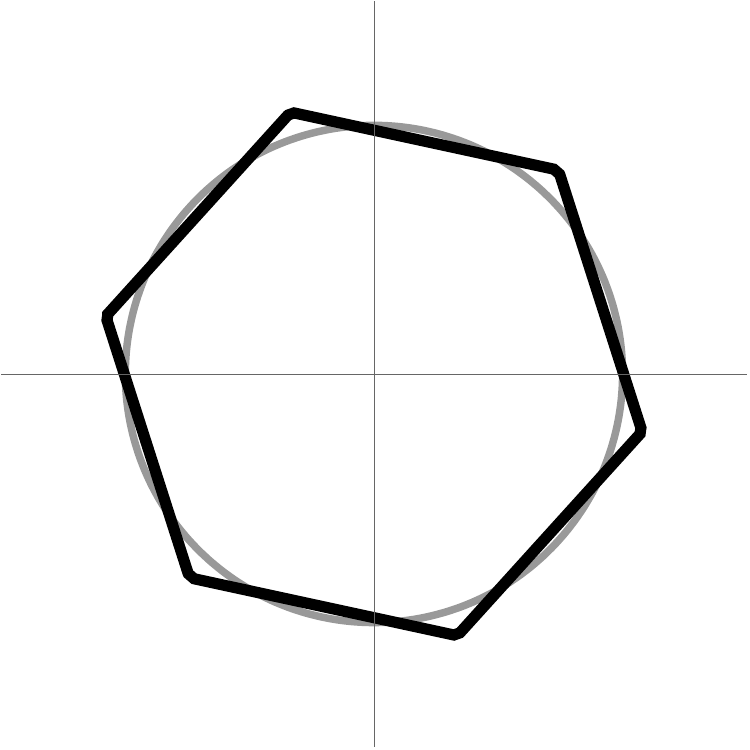}
\includegraphics[width=0.3\columnwidth]{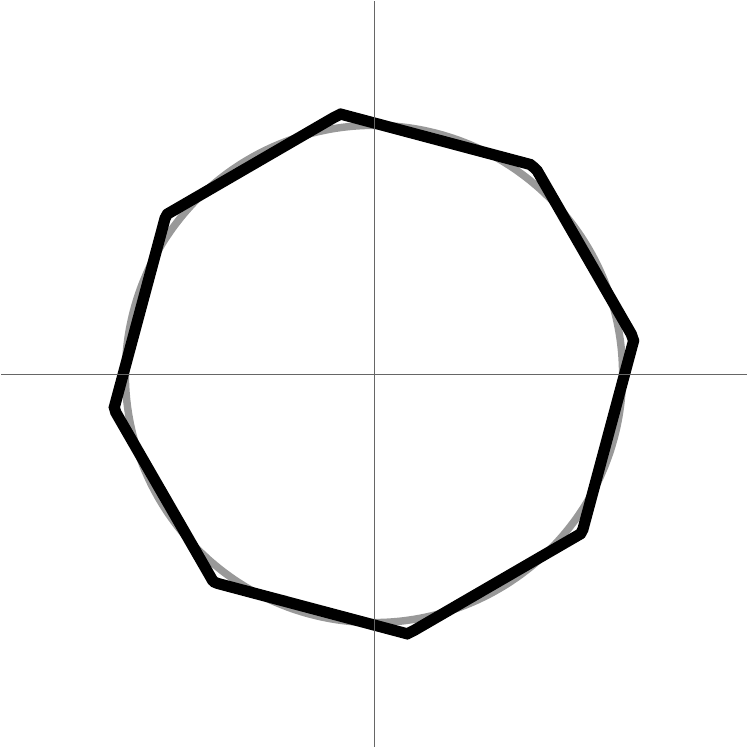}
\\
\includegraphics[width=0.3\columnwidth]{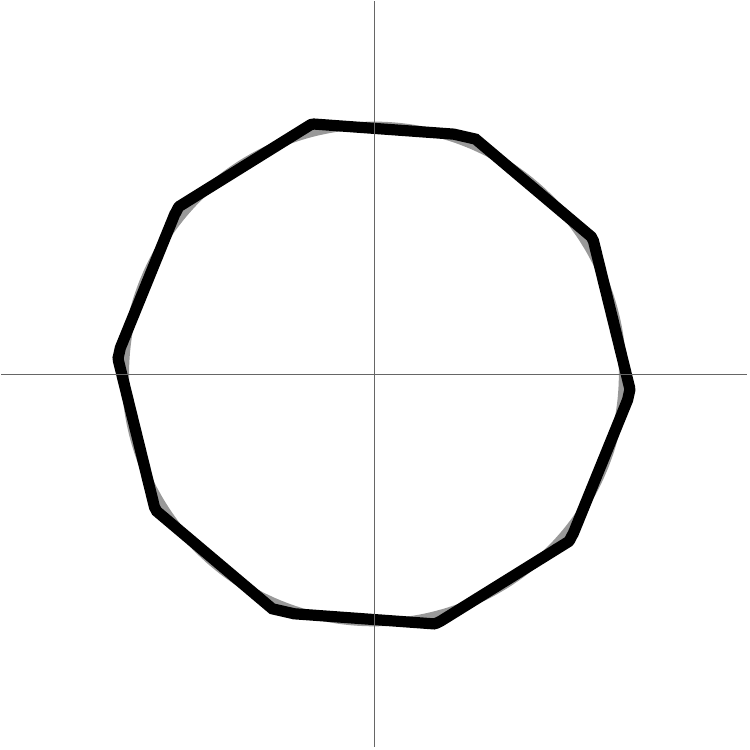}
\includegraphics[width=0.3\columnwidth]{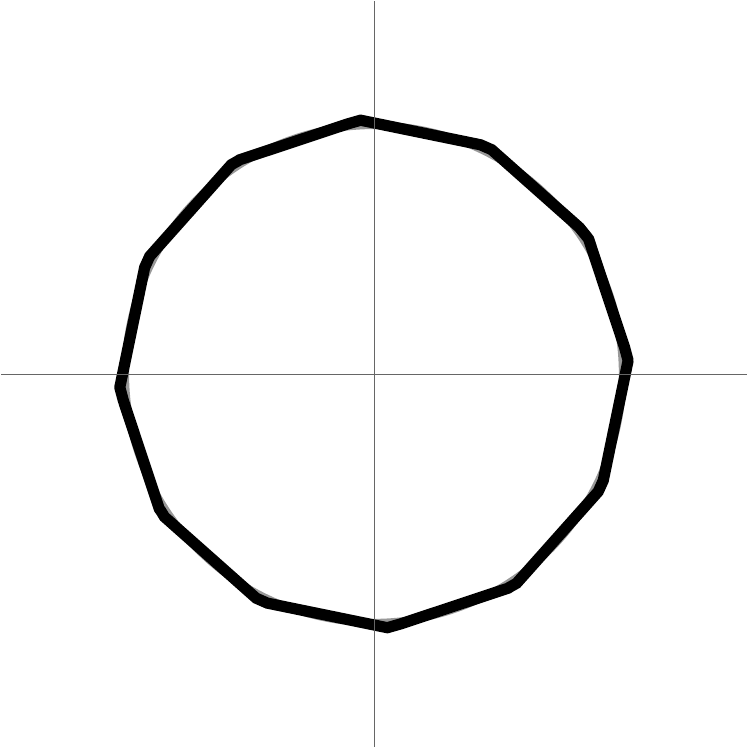}
\includegraphics[width=0.3\columnwidth]{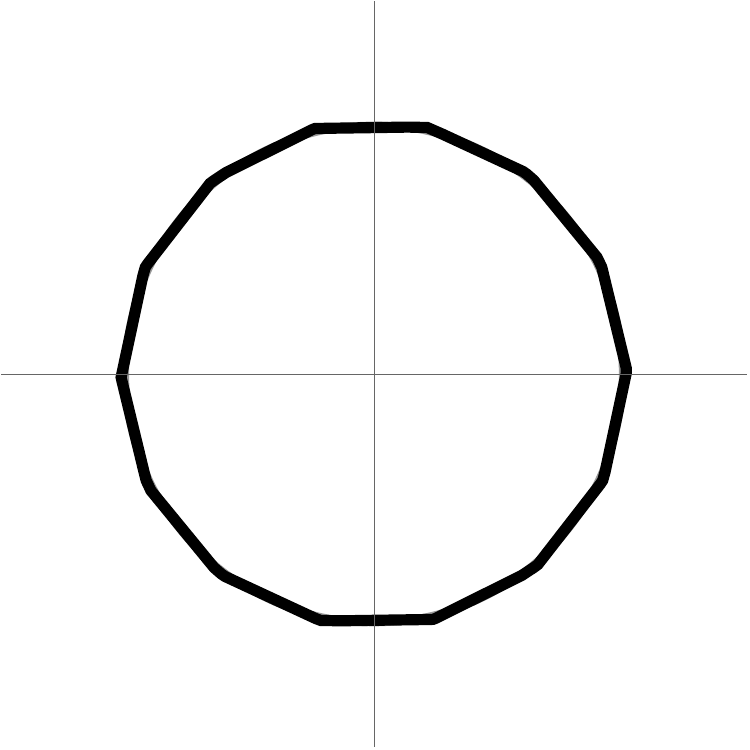}
\vspace*{-5mm}
\caption{Neural $2$-polytope (polygon) of type $(n;1)$ (the one generated by the network architecture with a single middle layer with the unit number $n$, and the activation function is $ \left| x \right|^1$), where $n=2$, $3$, $4$, $5$, $6$, $7$ (thick black lines, from the top-left to the bottom-right). These turn out to be $2n$-sided regular polygons. Grey lines are the unit circle used for the training.}
\label{fig1}
\end{center}
\vskip -0.2in
\end{figure}

\begin{figure}[t]
\begin{center}
\includegraphics[width=0.3\columnwidth]{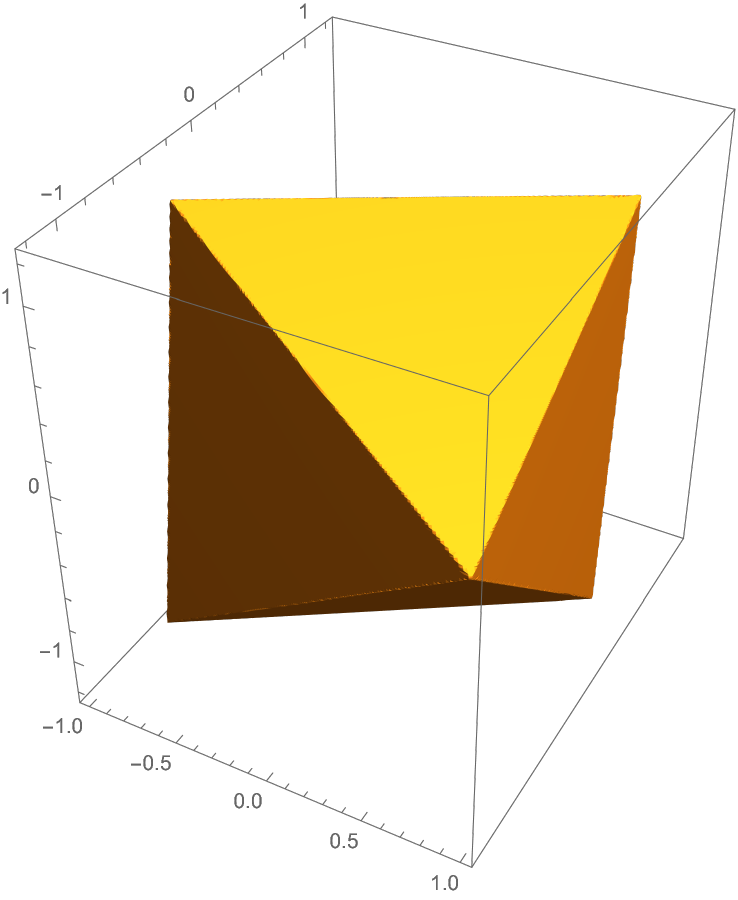}
\includegraphics[width=0.3\columnwidth]{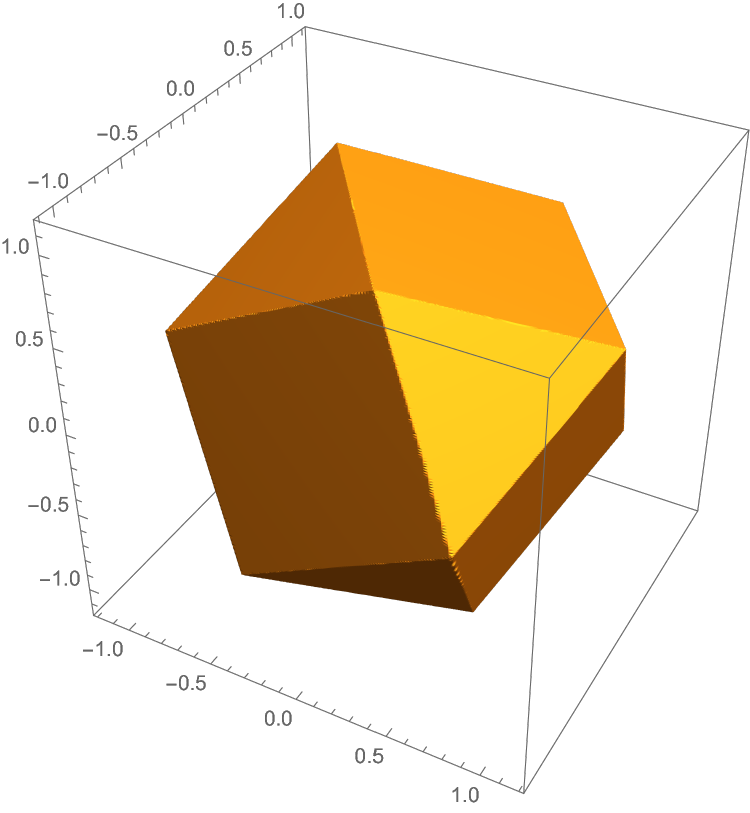}
\includegraphics[width=0.3\columnwidth]{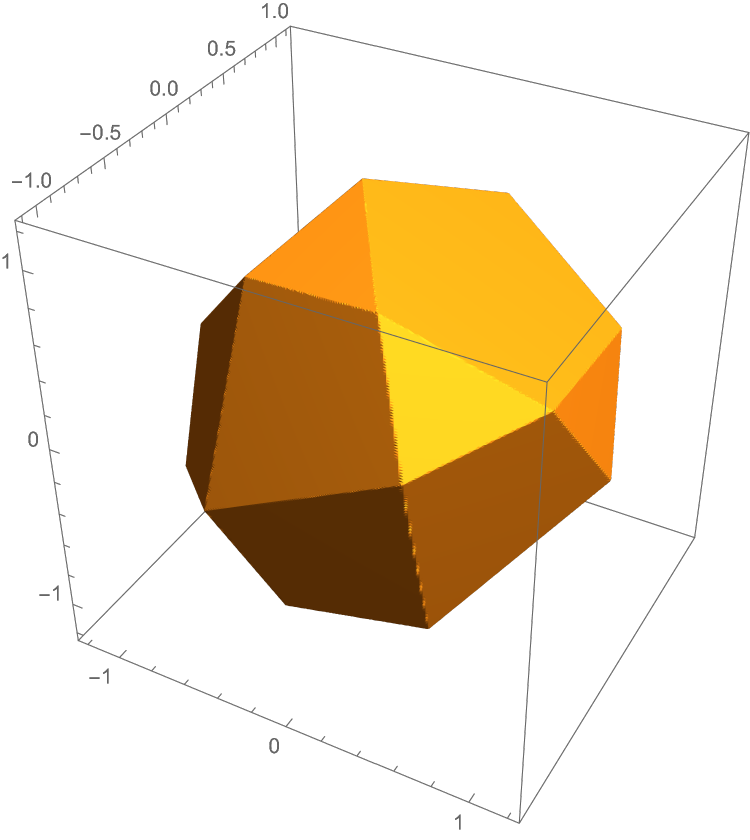}
\\
\includegraphics[width=0.3\columnwidth]{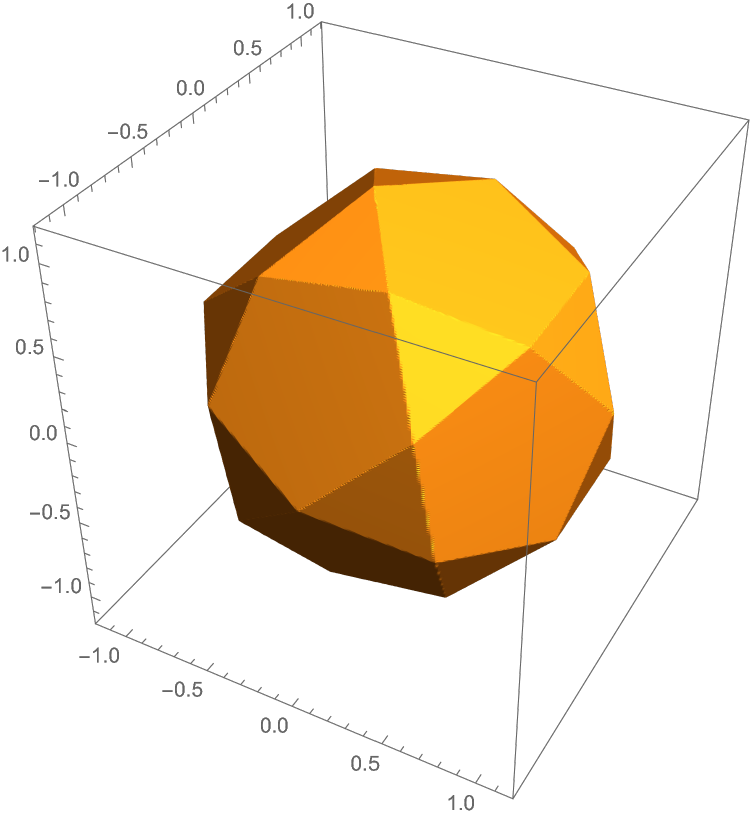}
\includegraphics[width=0.3\columnwidth]{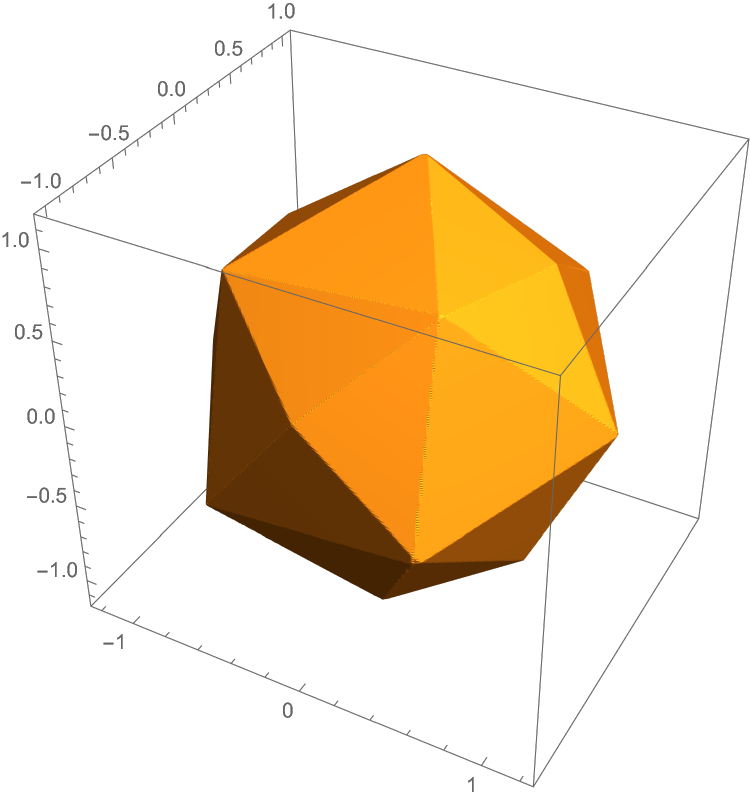}
\includegraphics[width=0.3\columnwidth]{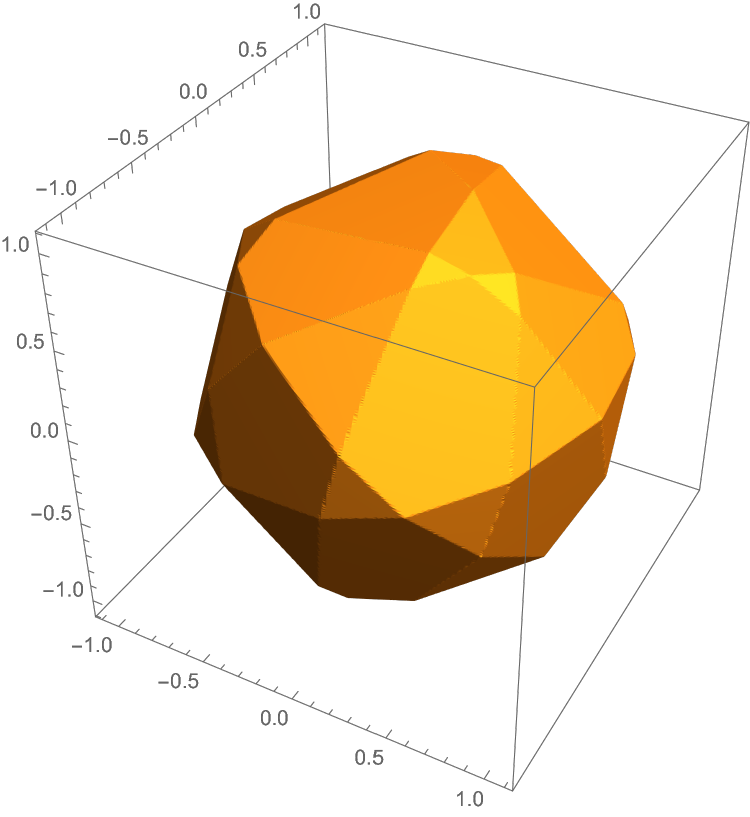}
\\
\includegraphics[width=0.23\columnwidth]{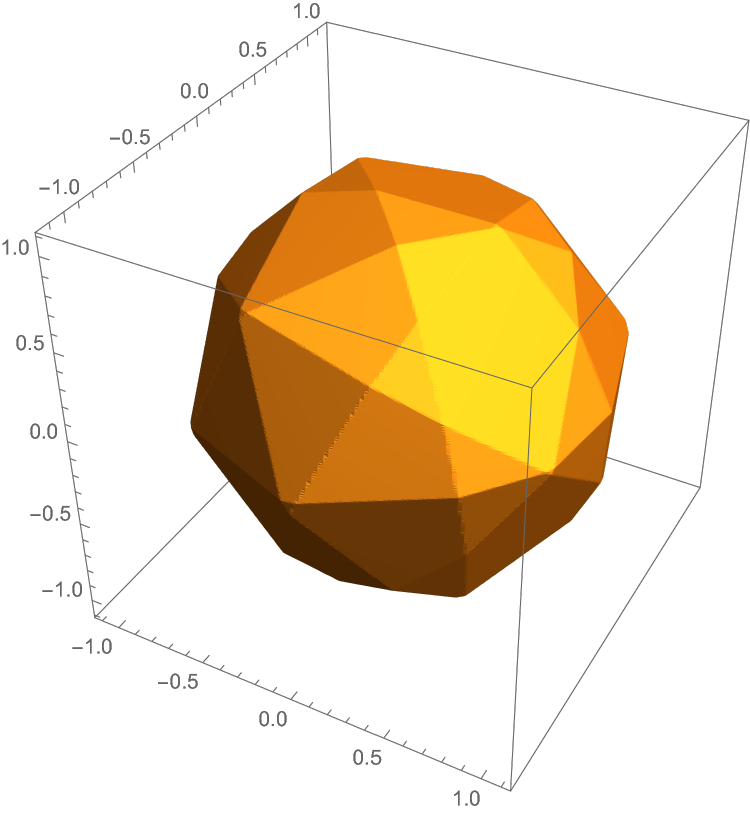}
\includegraphics[width=0.23\columnwidth]{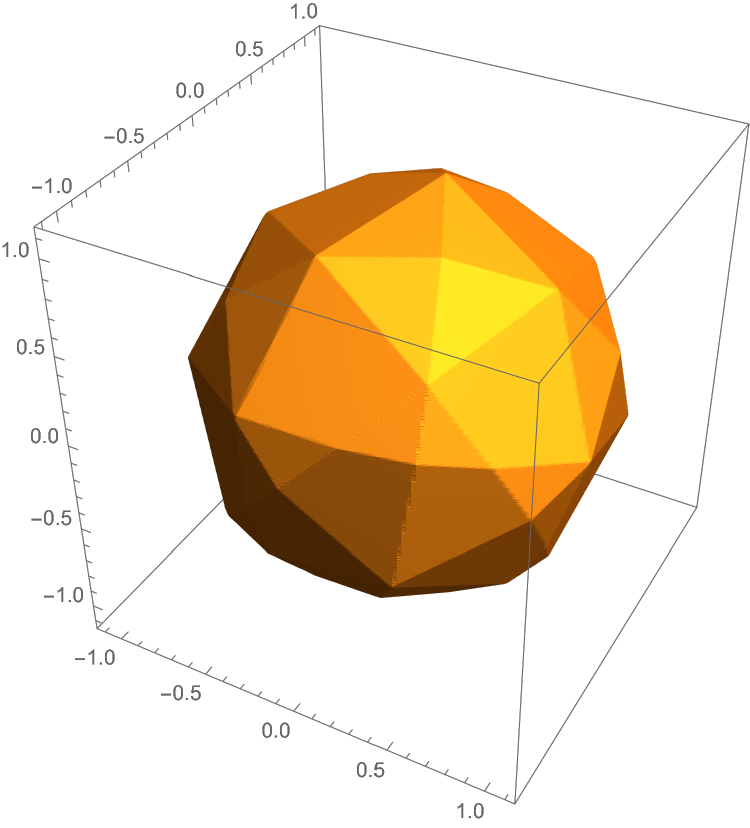}
\includegraphics[width=0.23\columnwidth]{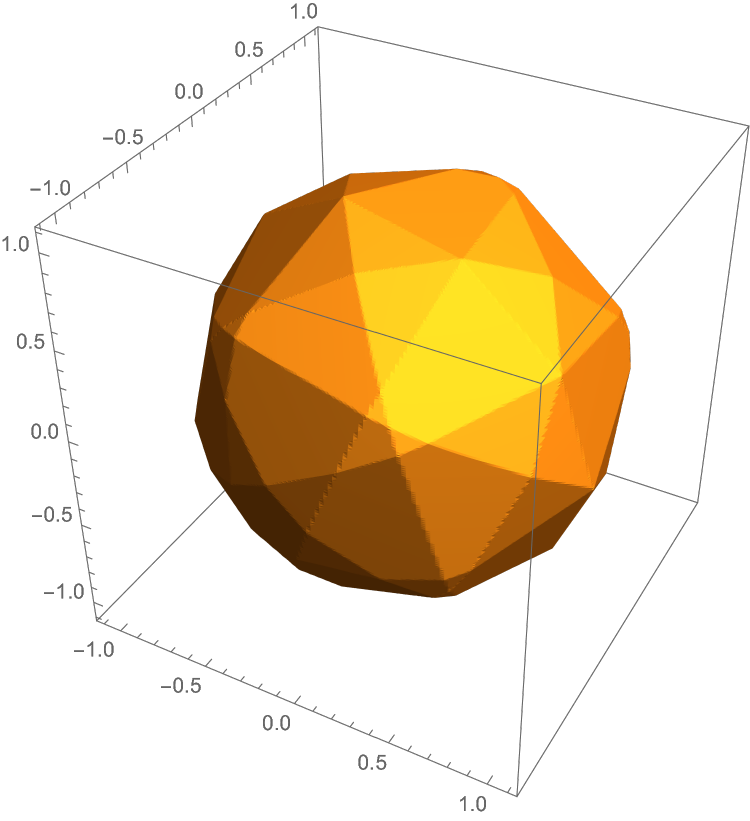}
\includegraphics[width=0.23\columnwidth]{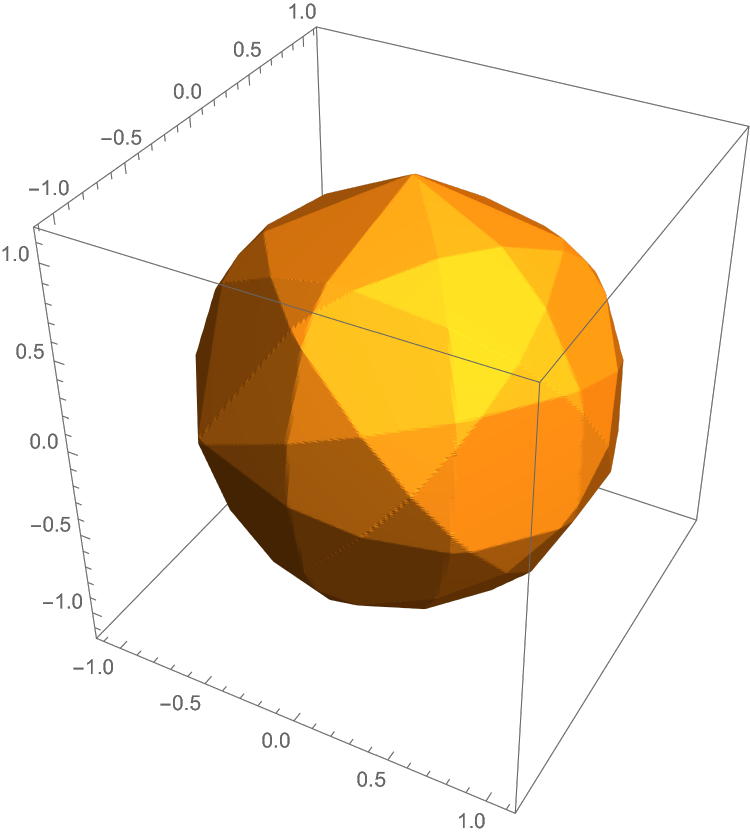}
\vspace*{-5mm}
\caption{Neural $3$-polytope (polyhedron) of type $(n;1)$ (the one generated by the network architecture with a single middle layer with the unit number $n$), where $n=3$, $4$, \ldots, $12$ (from the top-left to the bottom-right).}
\label{fig2}
\end{center}
\vskip -0.2in
\end{figure}

For the training, we produce roughly 10000 random points on the sphere, and use the ADAM optimizer with batch size 1000. 10000 epochs are enough for the training, as in this study we focus on very small architecture to see the discreteness of the neural network functions.\footnote{We do not show the explicit values of the loss functions after the training, as our neural network architecture is quite small in size and the minimum of the loss function landscape is expected to be unique, in all of our examples, up to the trivial flat directions generated by the rotation.}

After the training, we plot the cross section defined by
\begin{align}
    f(x_i) = 1,
    \label{eqfxi1}
\end{align}
where $f(x_i)$ is the trained neural network function. We call this cross section ``\textit{neural polytopes}.''
We name the produced polytopes as $d$-polytope of type $(n_1, \ldots, n_N; p_1, \ldots, p_N)$, where $d$ is the spatial dimension of the minimal Euclidean space in which the polytope is embedded (i.e. the number of input units), and $(n_1, \ldots, n_N)$ refers to the number of units in each of the $N$ intermediate layers, and $(p_1, \ldots, p_N)$ is the power appearing in the activation function $ \left| x \right|^{p_i}$ at each layer. When $p_1 = \cdots=p_N (=p)$, we just call it type $(n_1, \ldots, n_N; p)$. 




\begin{figure}[t]
\vskip 0.2in
\begin{center}
\includegraphics[width=0.2\columnwidth]{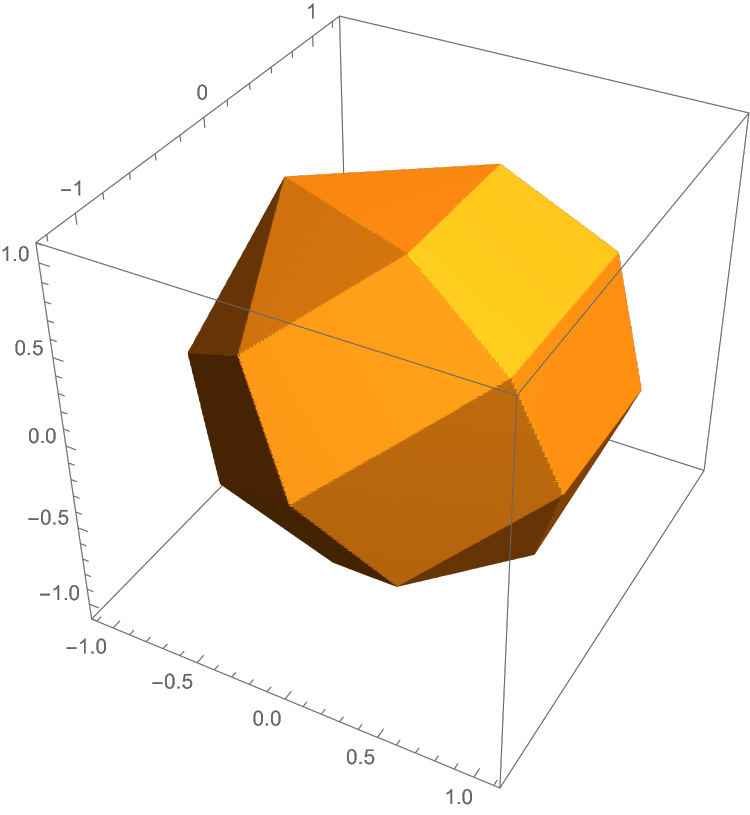}
\includegraphics[width=0.2\columnwidth]{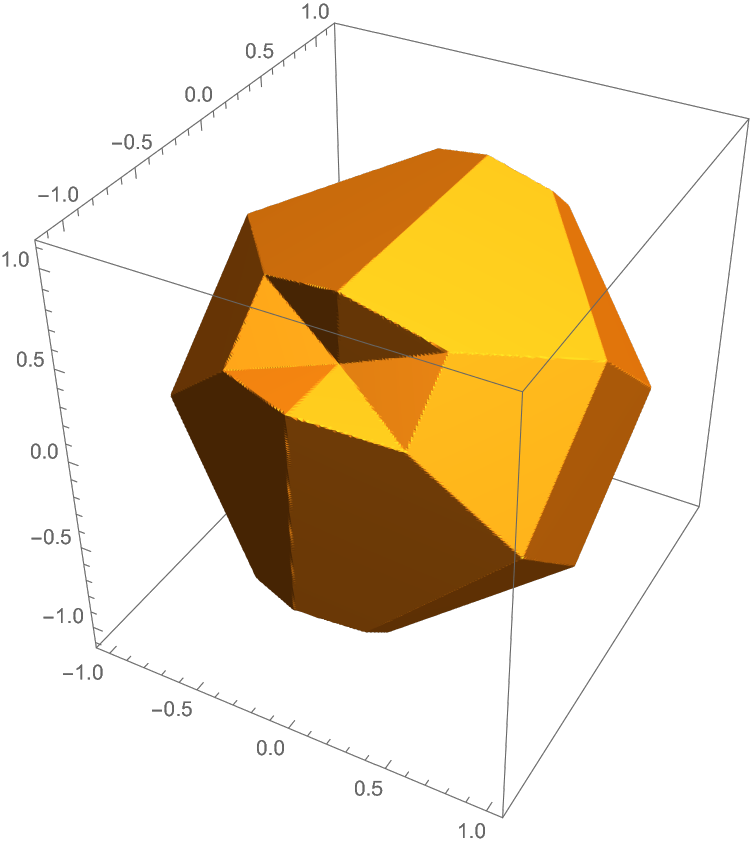}
\includegraphics[width=0.2\columnwidth]{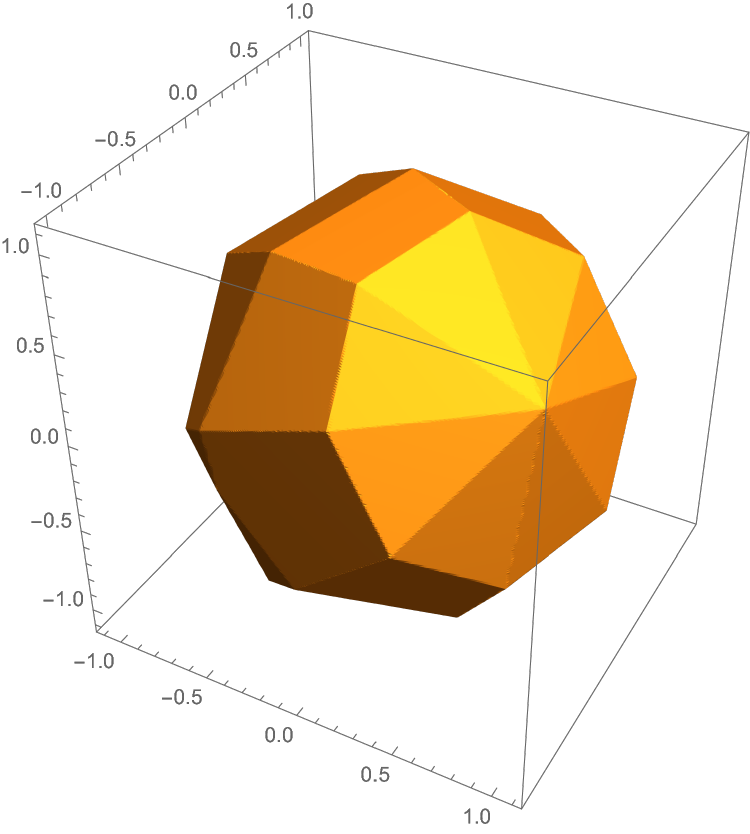}
\includegraphics[width=0.2\columnwidth]{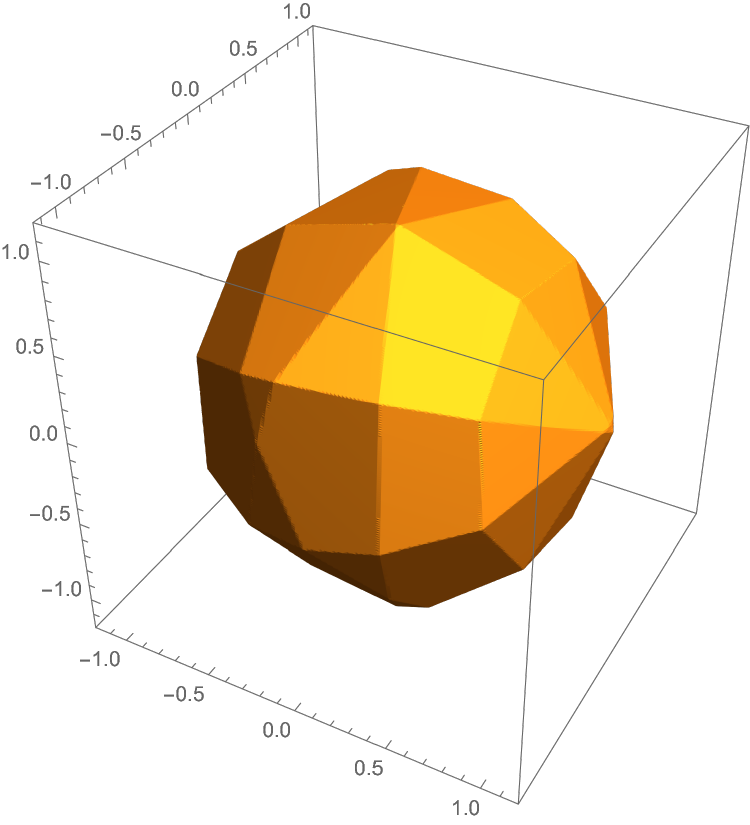}
\\
\includegraphics[width=0.2\columnwidth]{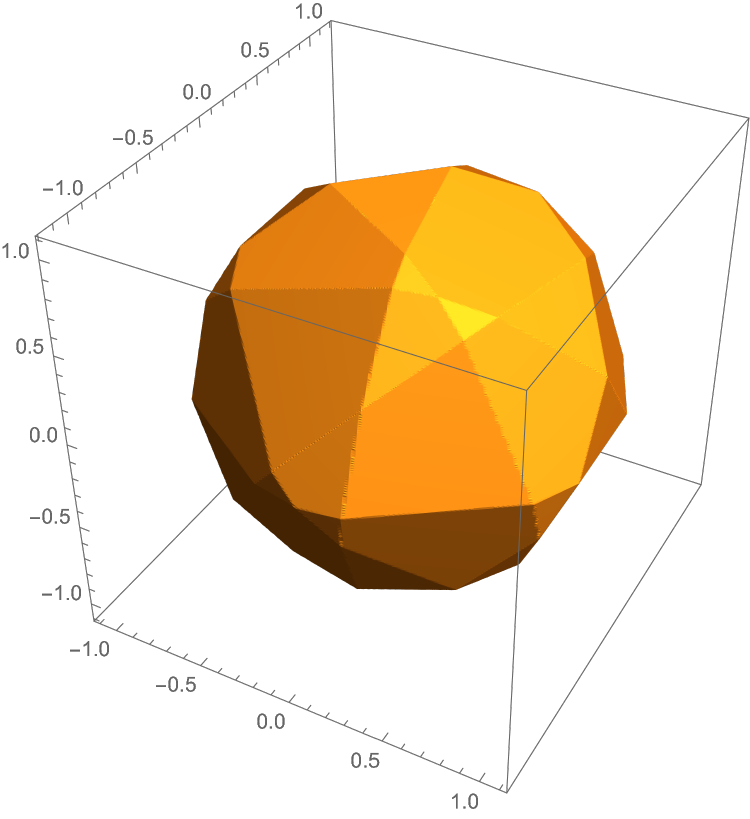}
\includegraphics[width=0.2\columnwidth]{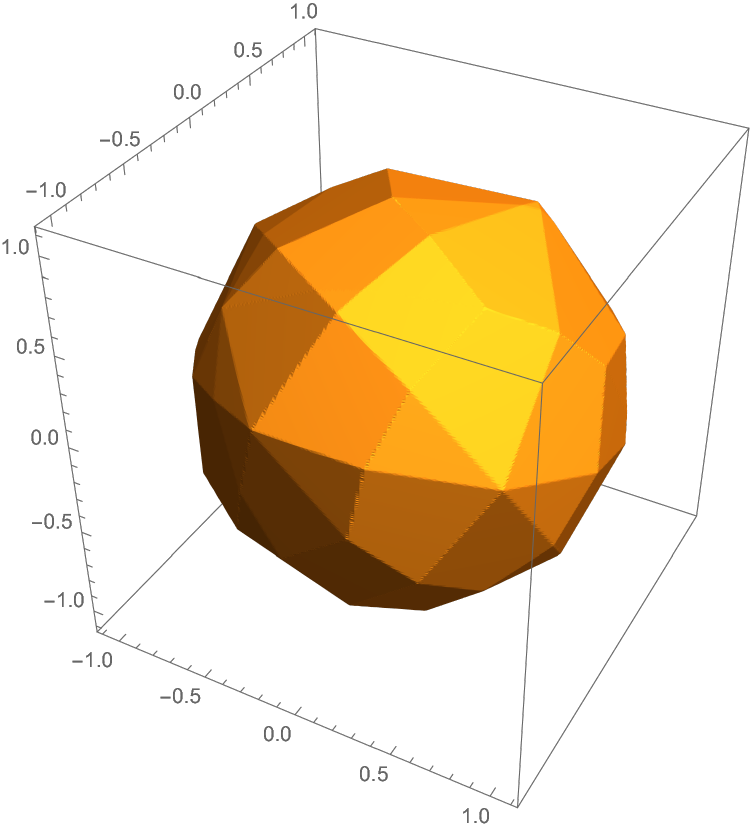}
\includegraphics[width=0.2\columnwidth]{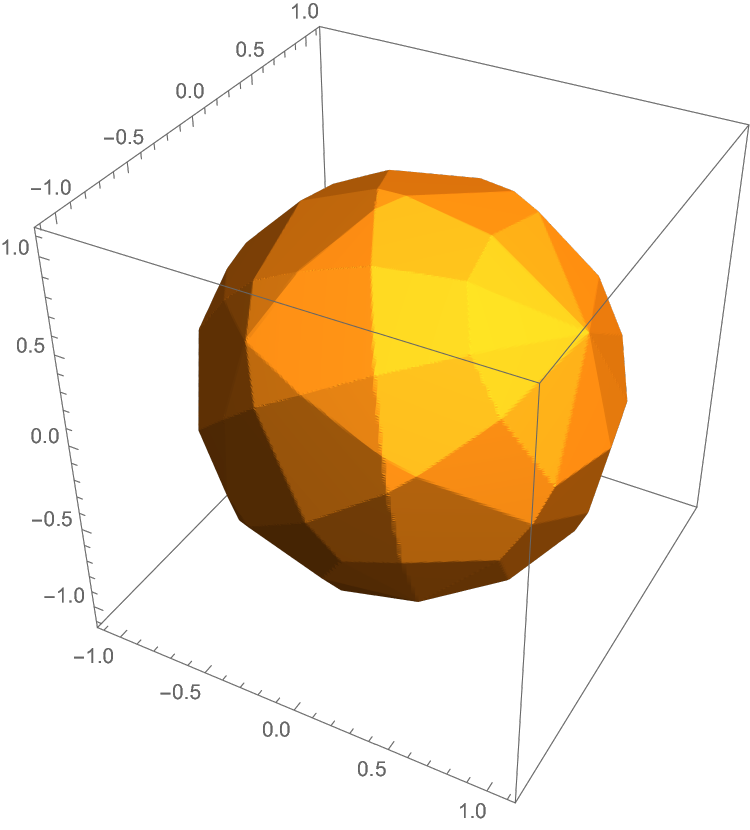}
\includegraphics[width=0.2\columnwidth]{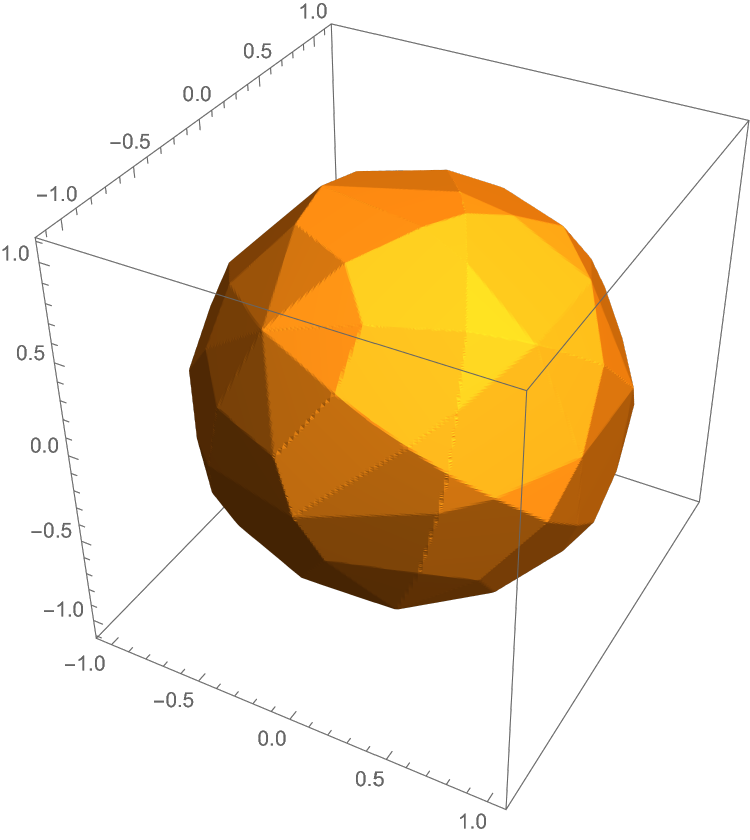}
\caption{Neurally generated polyhedra of the type $(n,2;1)$ (the ones generated by the network architecture  with two middle layers with the unit number $n$ and 2 for each layer), where $n=4$, \ldots, $11$ (from the left to the right).}
\label{fig3}
\end{center}
\vskip -0.2in
\end{figure}

\begin{figure*}[!ht]
\vskip 0.2in
\begin{center}
\centerline{
\includegraphics[width=0.2\columnwidth]{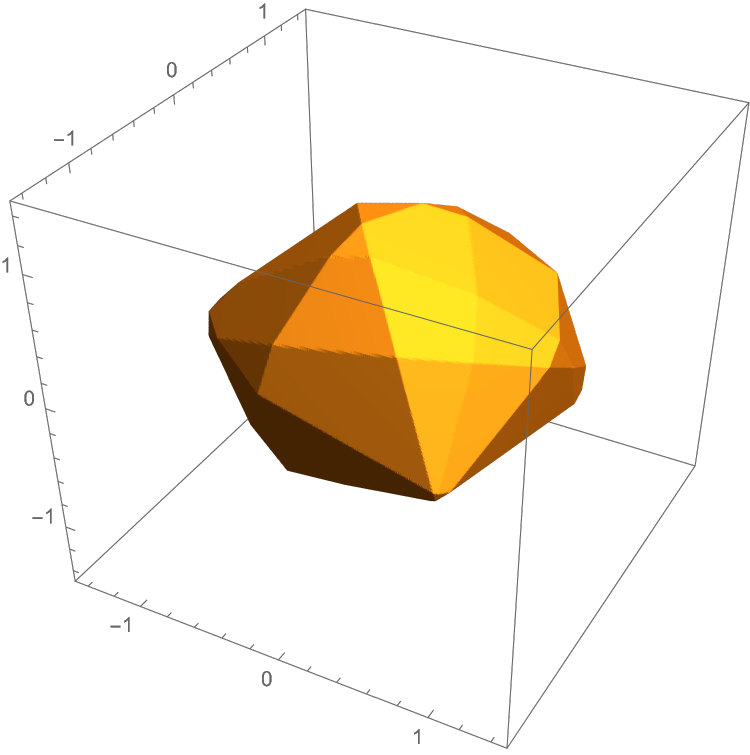}
\includegraphics[width=0.2\columnwidth]{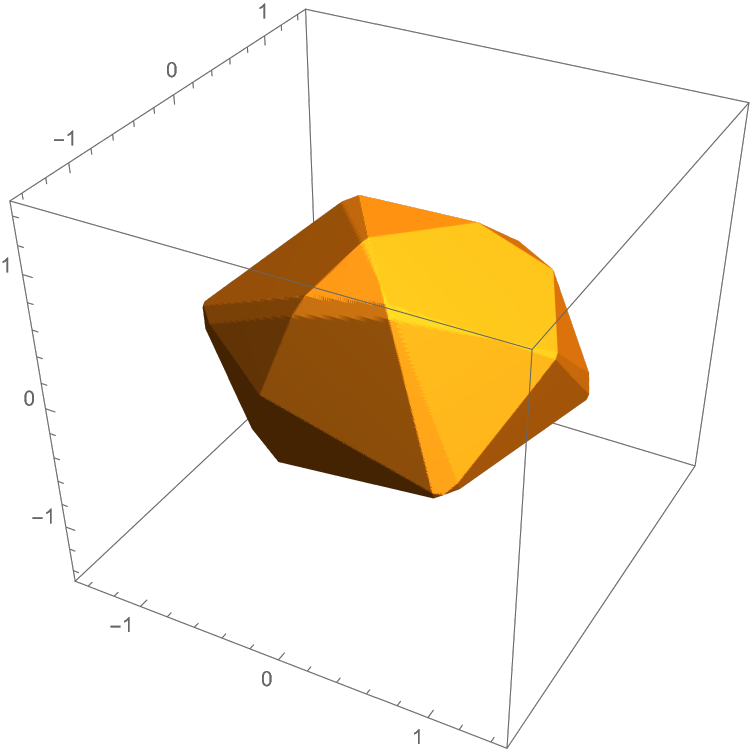}
\includegraphics[width=0.2\columnwidth]{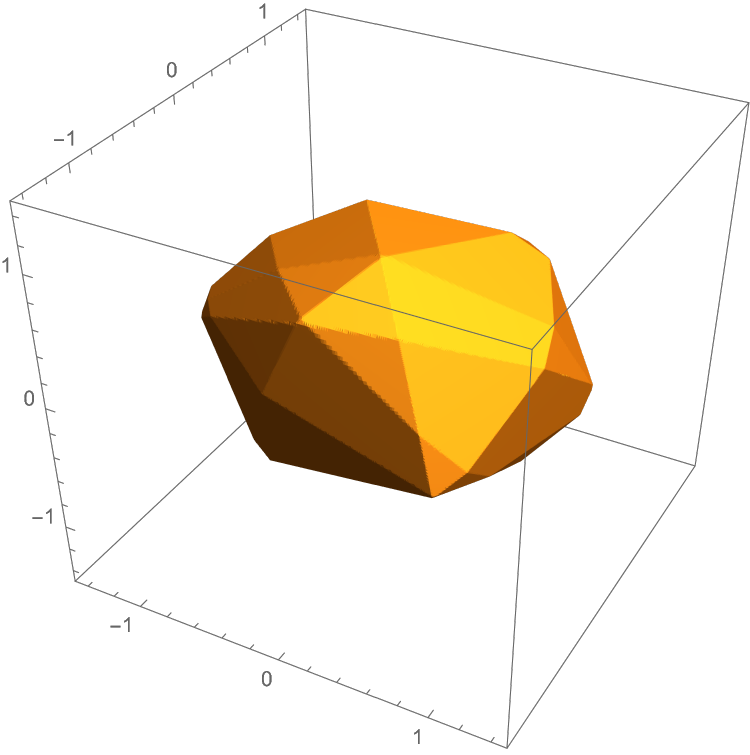}
\includegraphics[width=0.2\columnwidth]{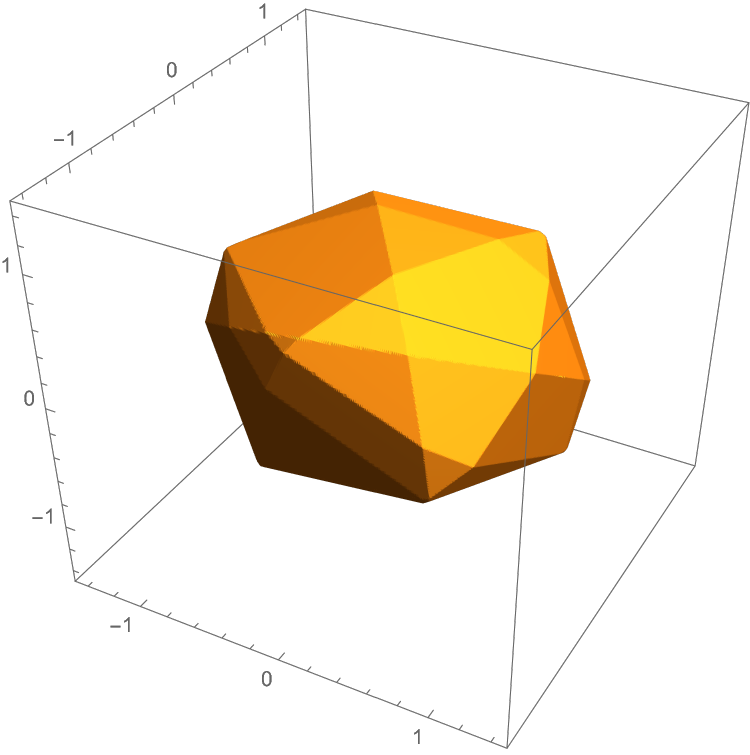}
\includegraphics[width=0.2\columnwidth]{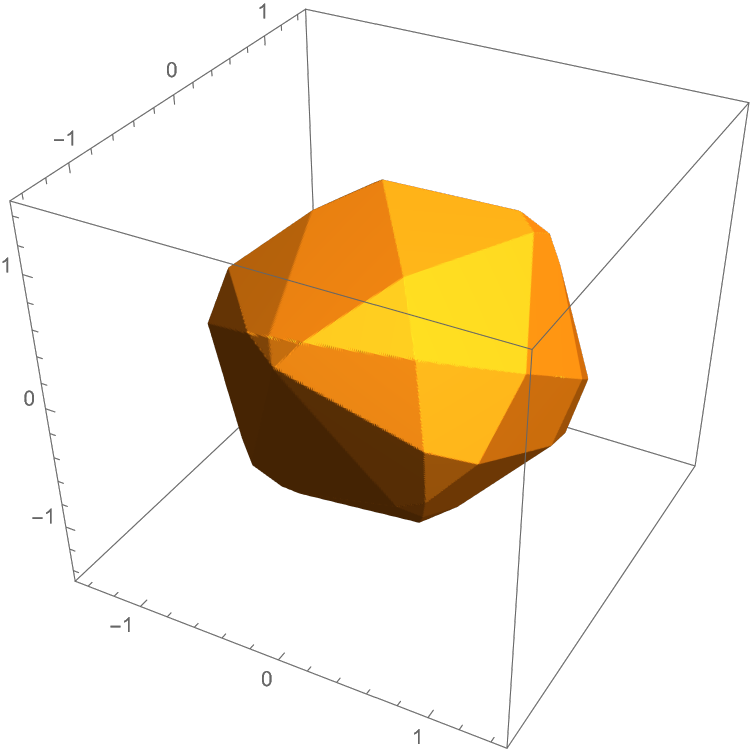}
\includegraphics[width=0.2\columnwidth]{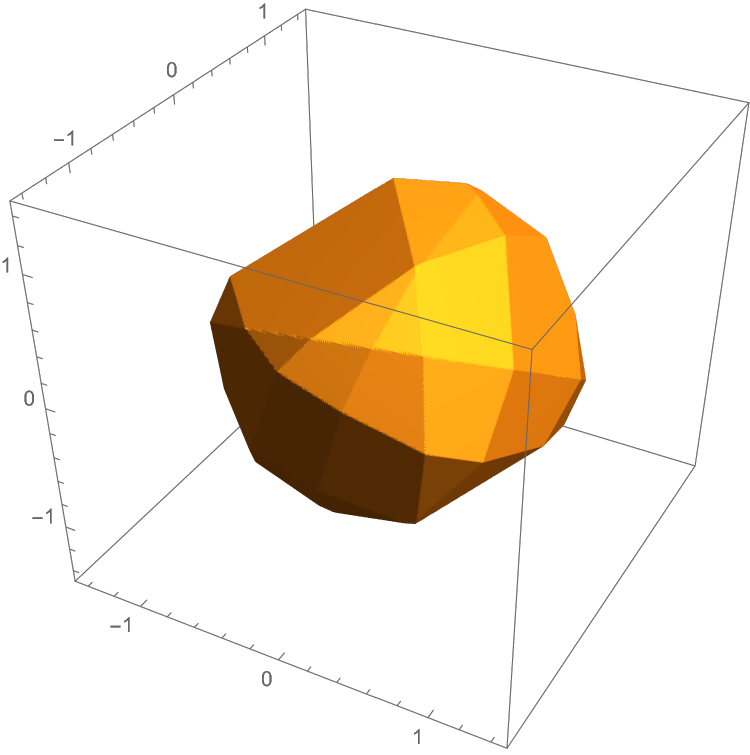}
\includegraphics[width=0.2\columnwidth]{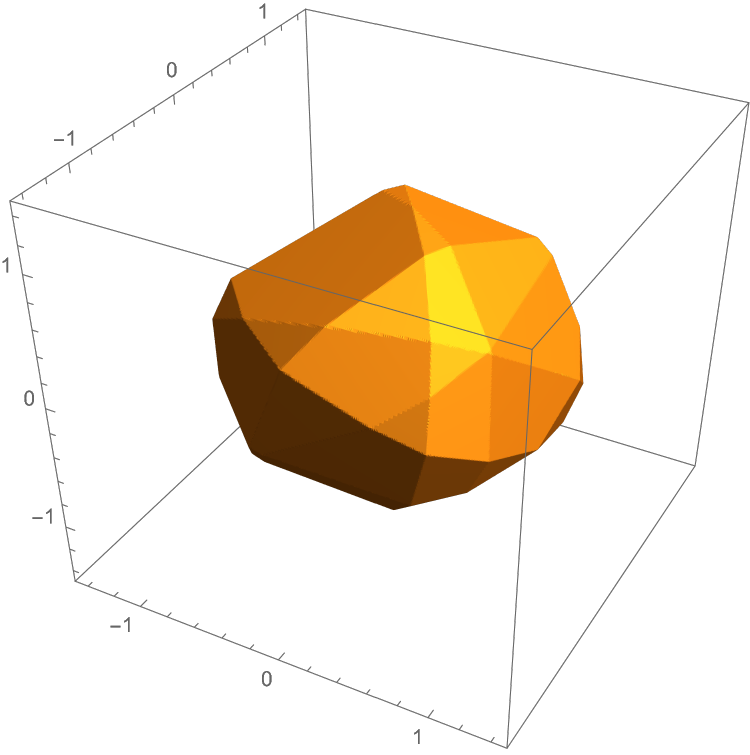}
\includegraphics[width=0.2\columnwidth]{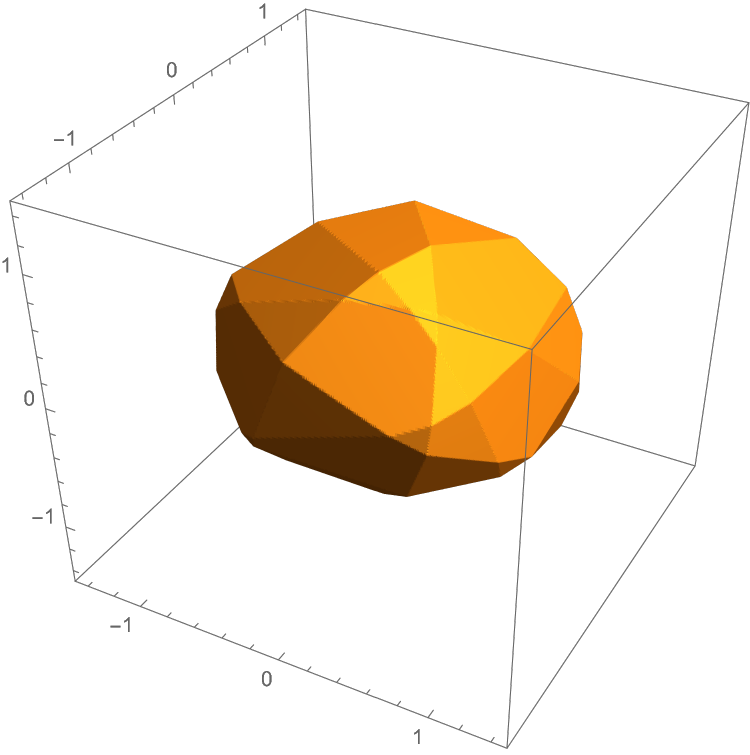}
\includegraphics[width=0.2\columnwidth]{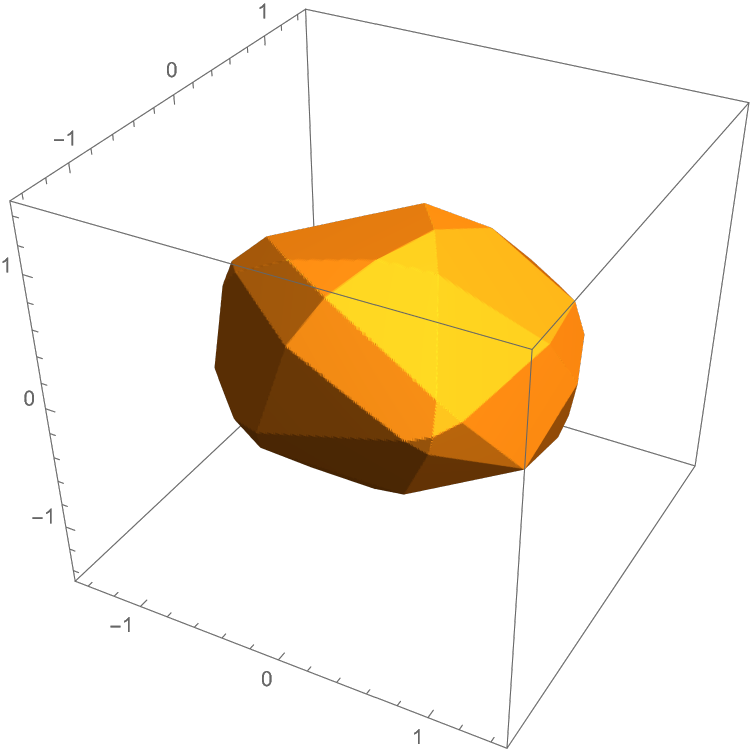}
\includegraphics[width=0.2\columnwidth]{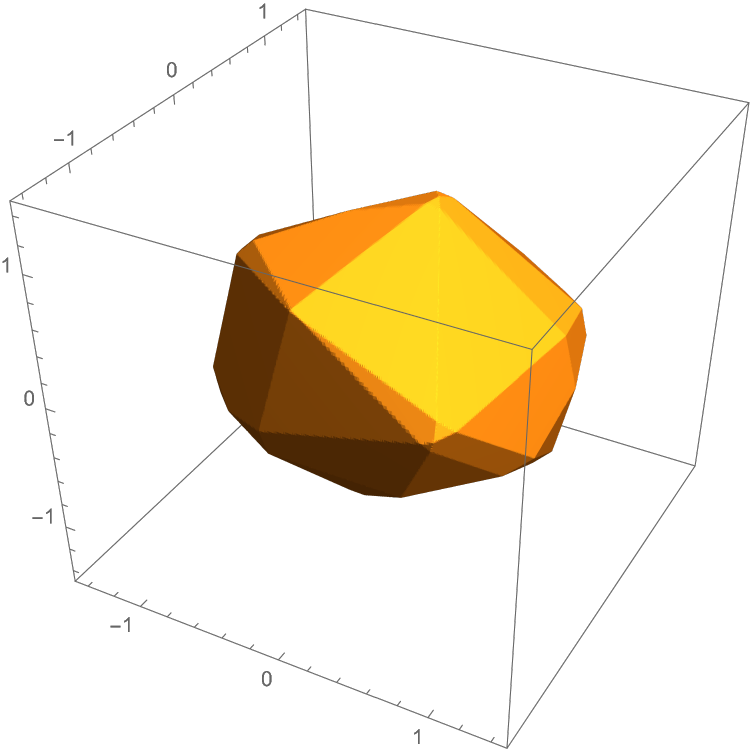}
}
\caption{Neural 5-polytopes of type $(8;1)$, with the activation function $|x|$, sliced at the plane $x_3=0$ and $x_4 = x_5 \tan \theta$, where $\theta = n \pi/10$ with $n=0$, $1$, $2$, $3$, \ldots, $9$ (from the left to the right).}
\label{fig4}
\end{center}
\end{figure*}

\begin{figure*}[!h]
\vskip 0.2in
\begin{center}
\centerline{
\includegraphics[width=0.2\columnwidth]{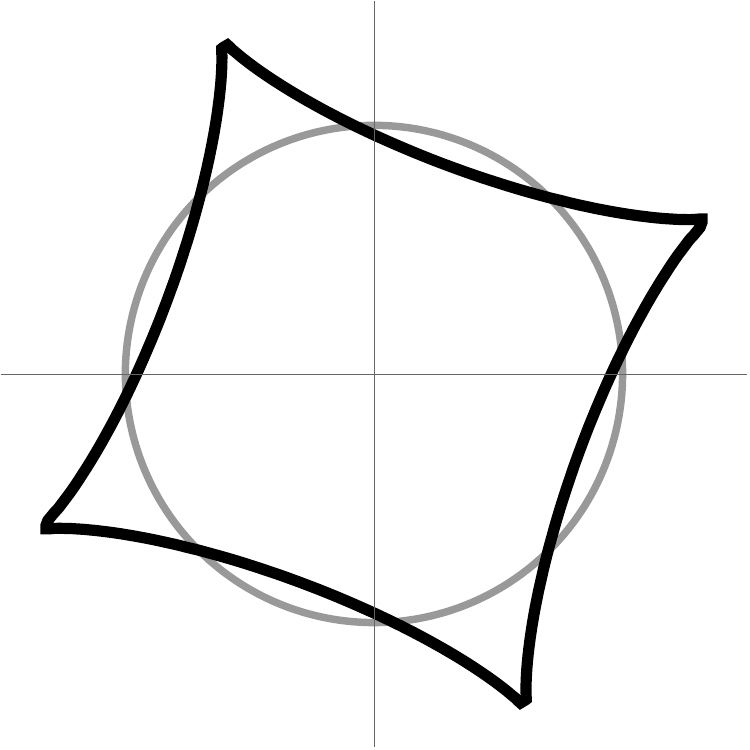}
\includegraphics[width=0.2\columnwidth]{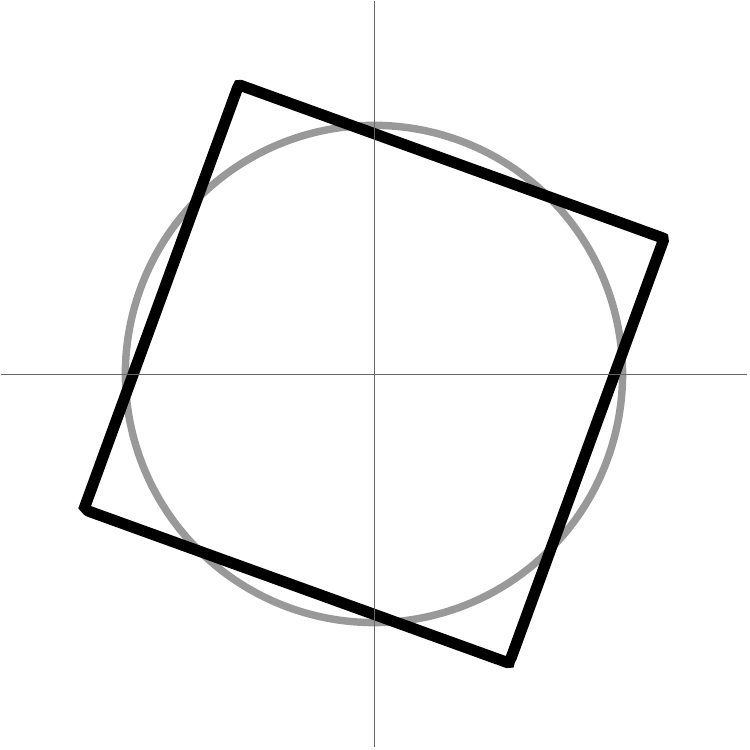}
\includegraphics[width=0.2\columnwidth]{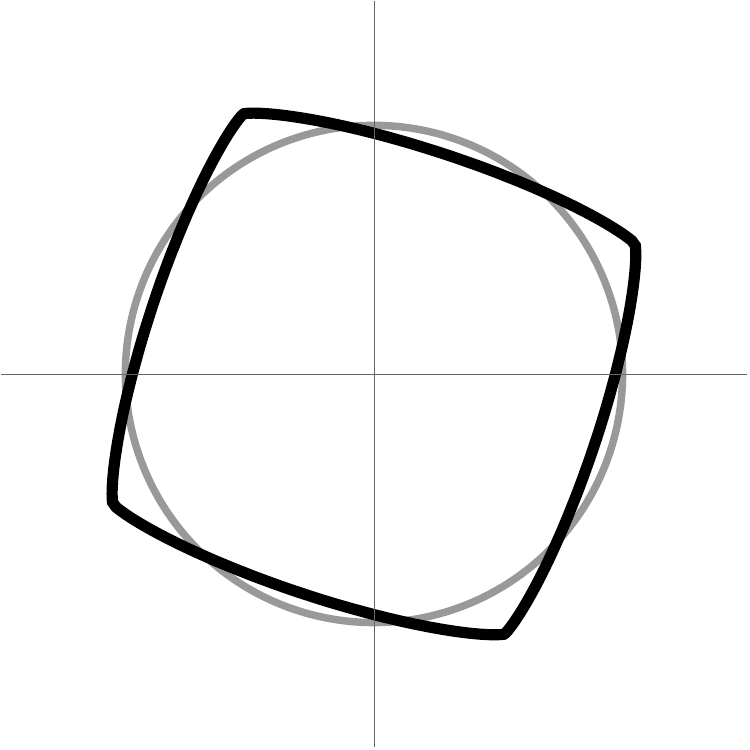}
\includegraphics[width=0.2\columnwidth]{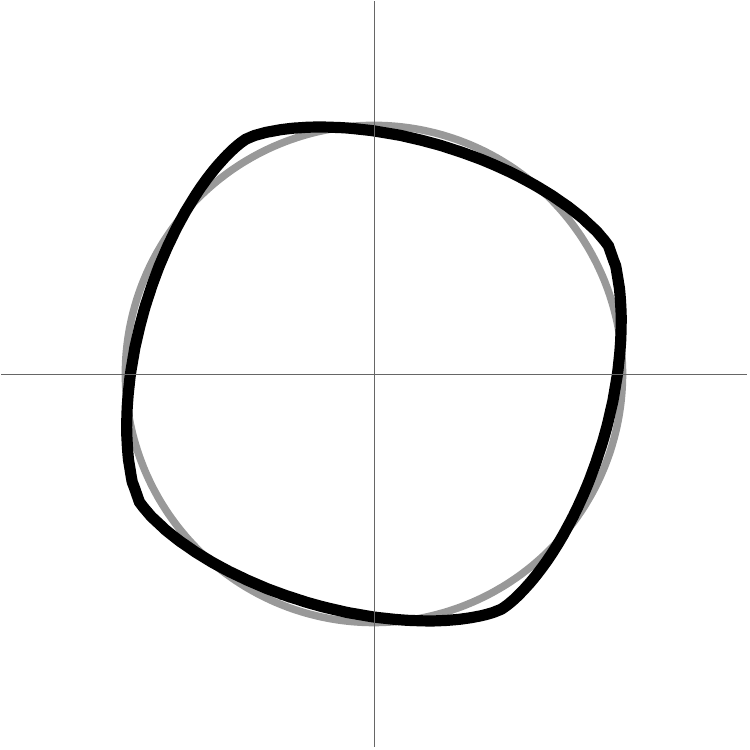}
\includegraphics[width=0.2\columnwidth]{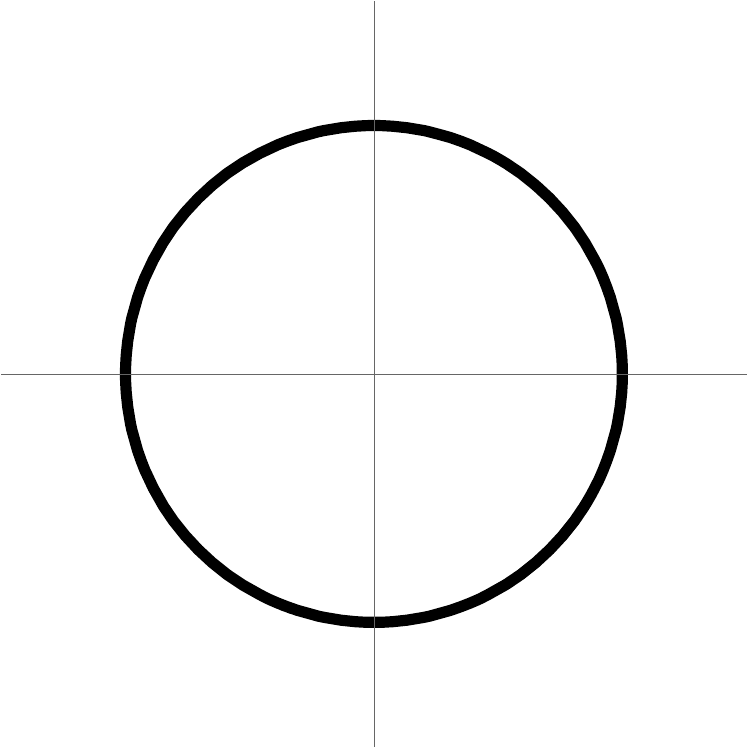}
\includegraphics[width=0.2\columnwidth]{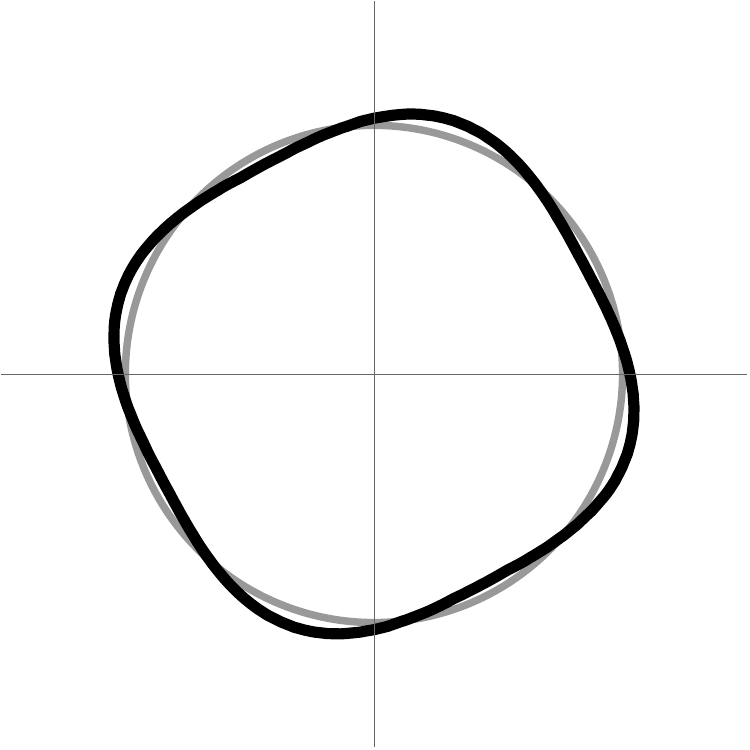}
\includegraphics[width=0.2\columnwidth]{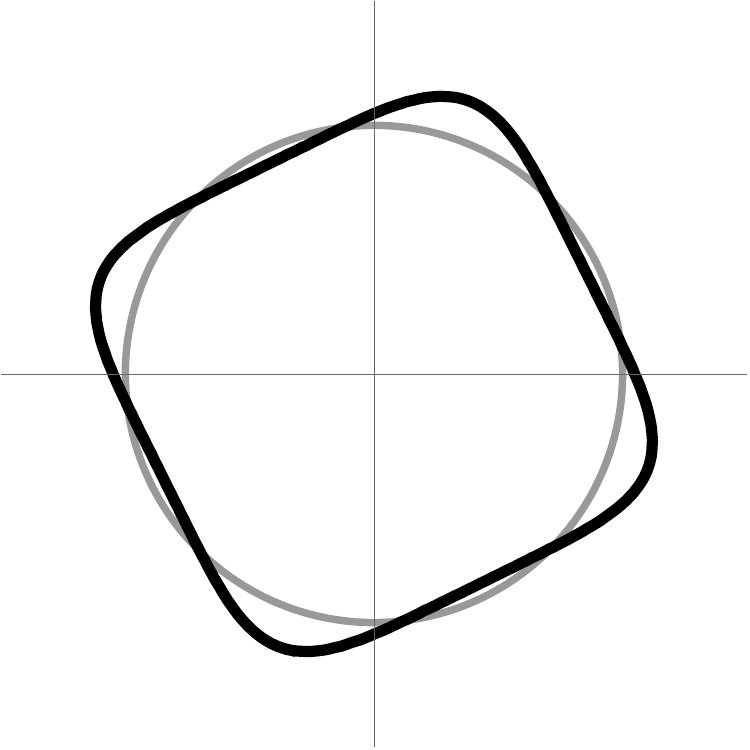}
\includegraphics[width=0.2\columnwidth]{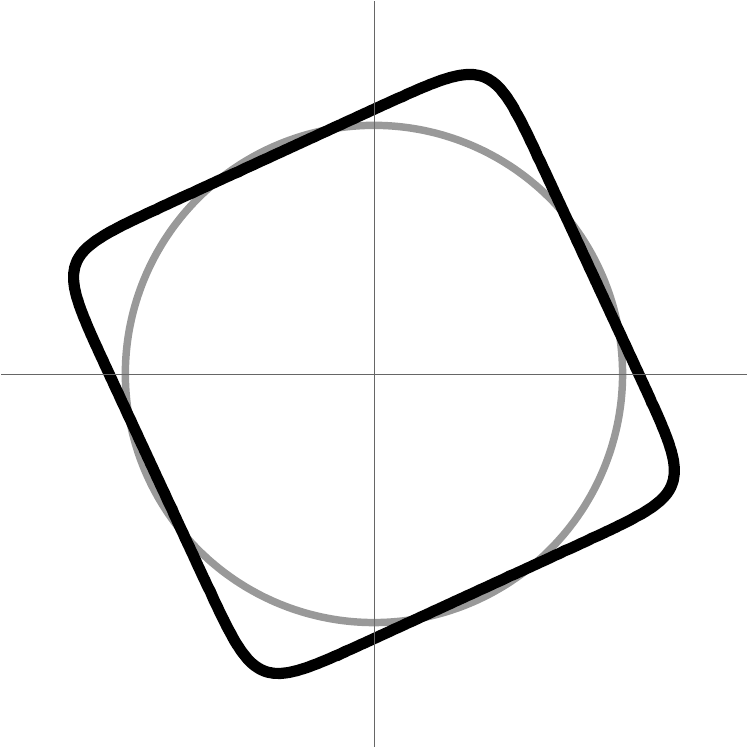}
}
\caption{Neural polygons of type $(2;p)$, with the activation function is chosen as $\left| x \right|^p$, where $p=0.8$, $1.0$, $1.2$, $1.5$, $2$, $3$, $5$, $10$ (from the left to the right).}
\label{fig5}
\end{center}
\end{figure*}

\begin{figure*}[!h]
\vskip 0.2in
\begin{center}
\centerline{
\includegraphics[width=0.2\columnwidth]{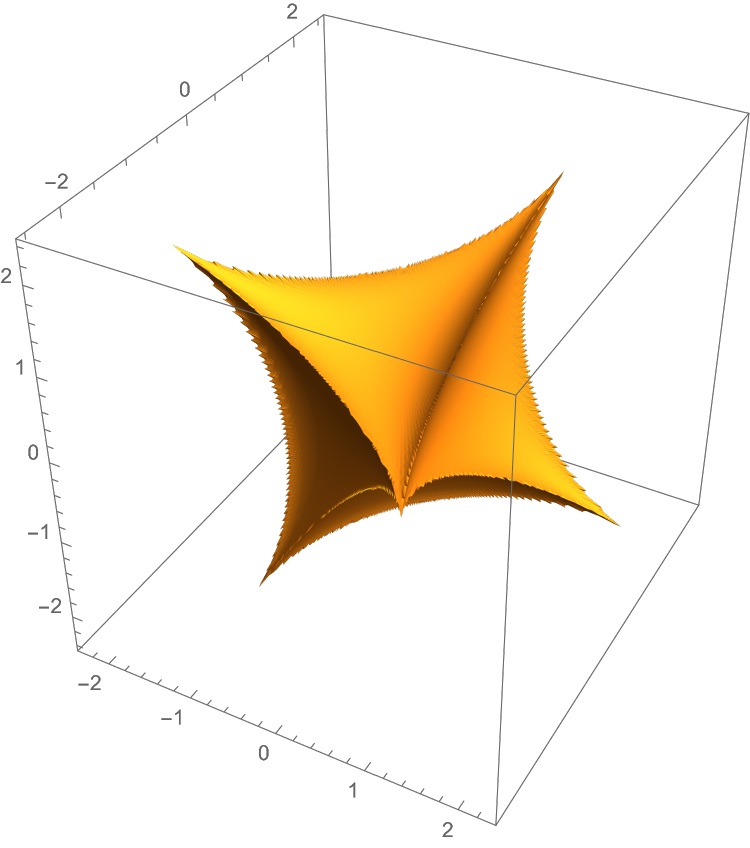}
\includegraphics[width=0.2\columnwidth]{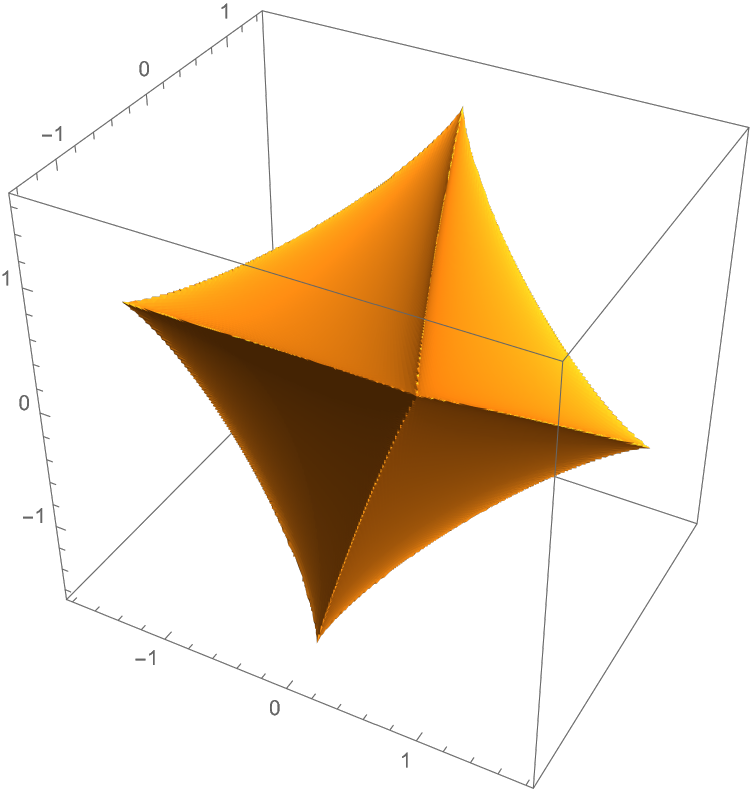}
\includegraphics[width=0.2\columnwidth]{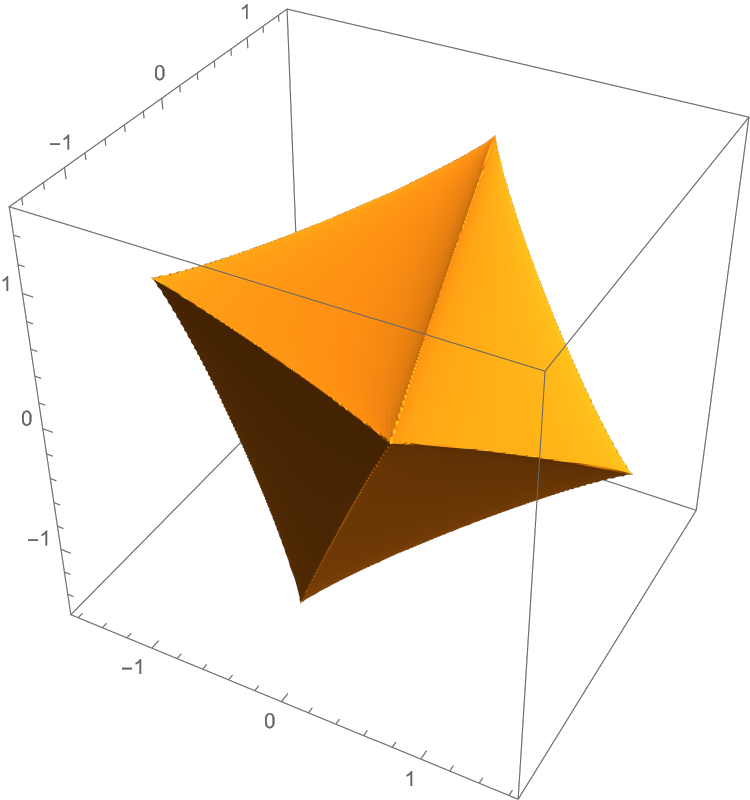}
\includegraphics[width=0.2\columnwidth]{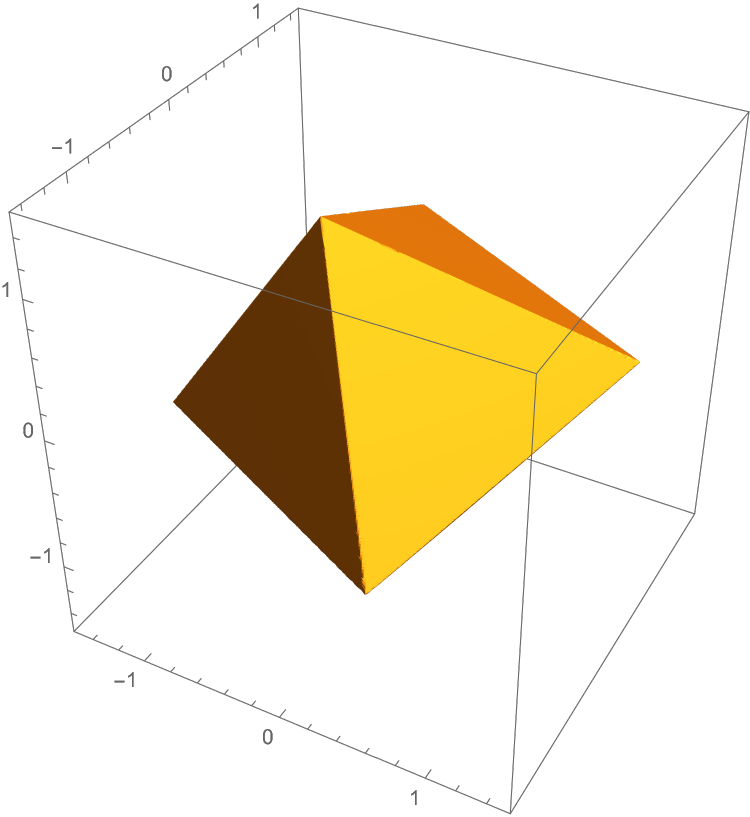}
\includegraphics[width=0.2\columnwidth]{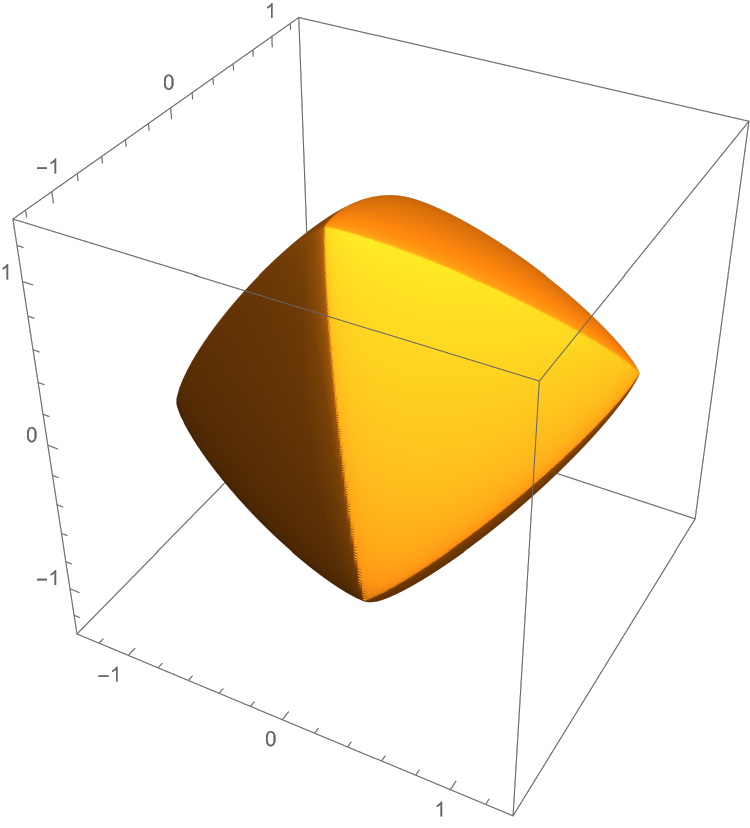}
\includegraphics[width=0.2\columnwidth]{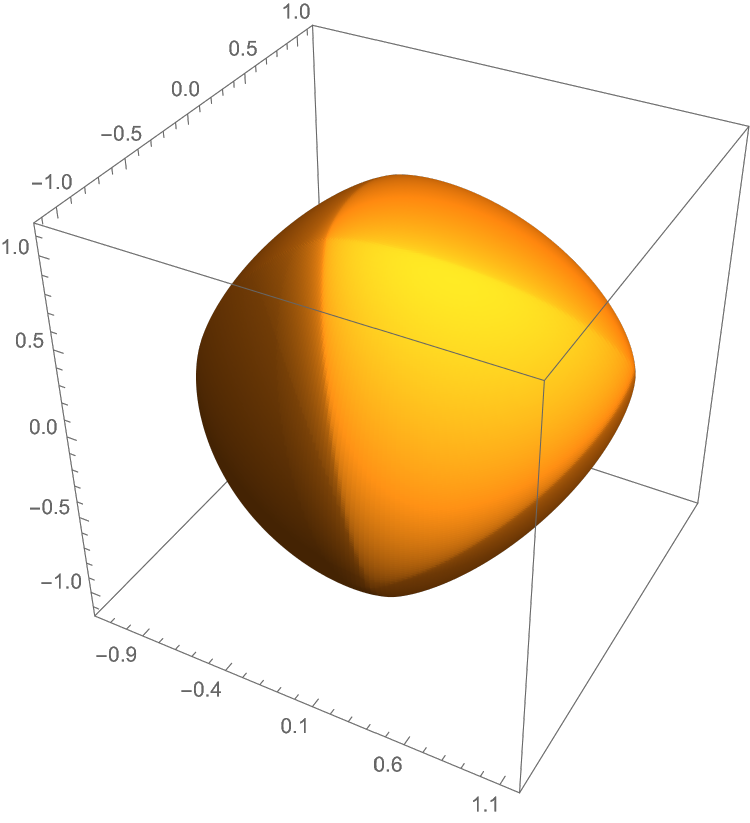}
\includegraphics[width=0.2\columnwidth]{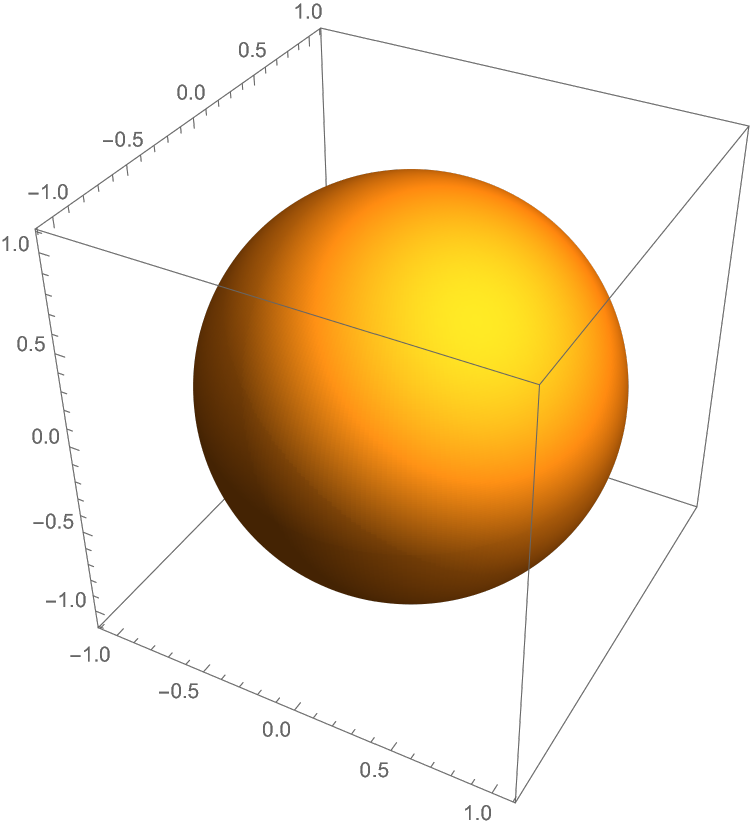}
\includegraphics[width=0.2\columnwidth]{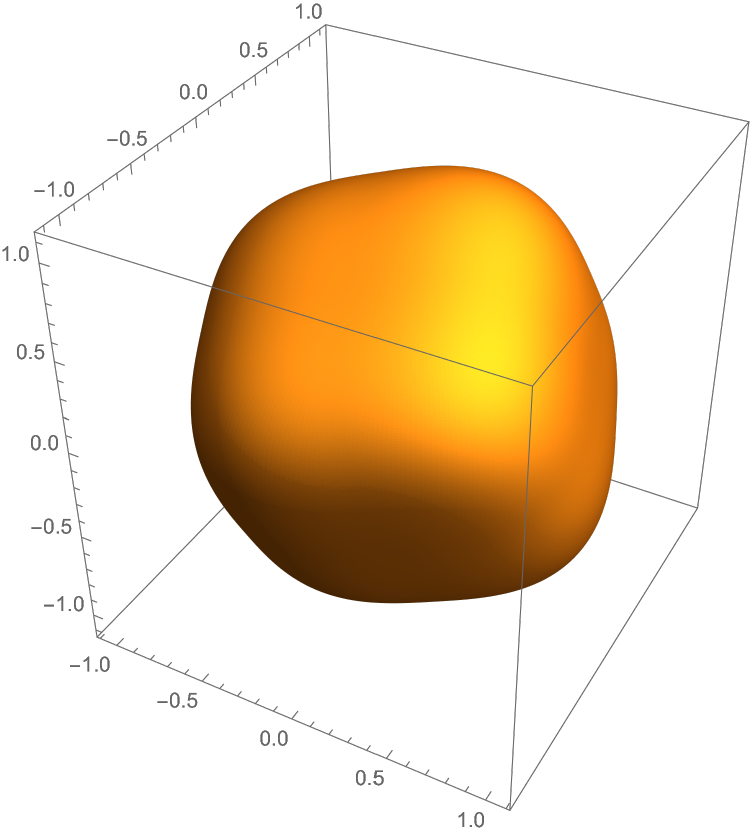}
\includegraphics[width=0.2\columnwidth]{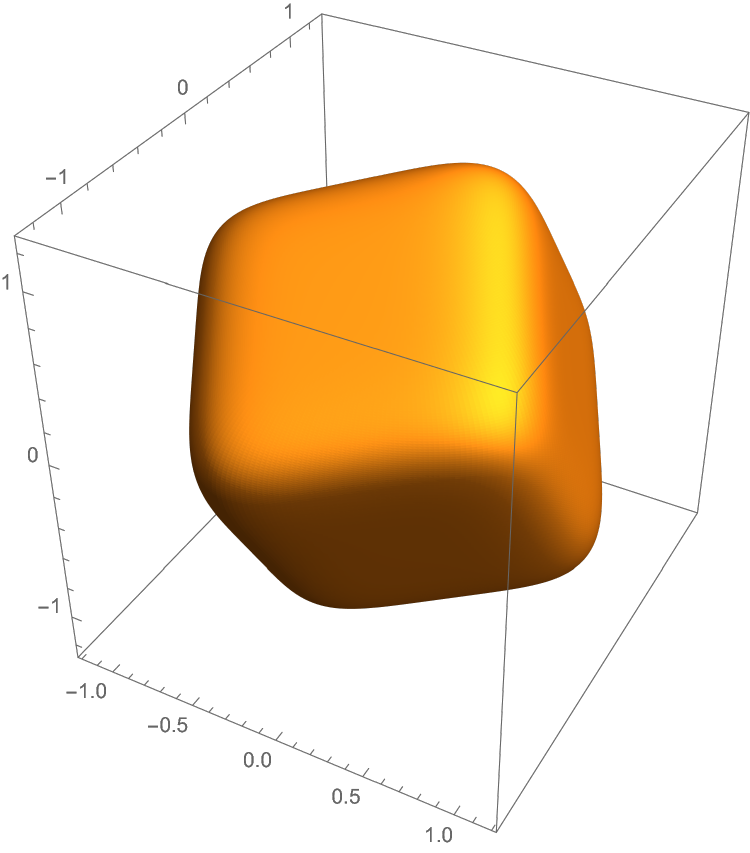}
\includegraphics[width=0.2\columnwidth]{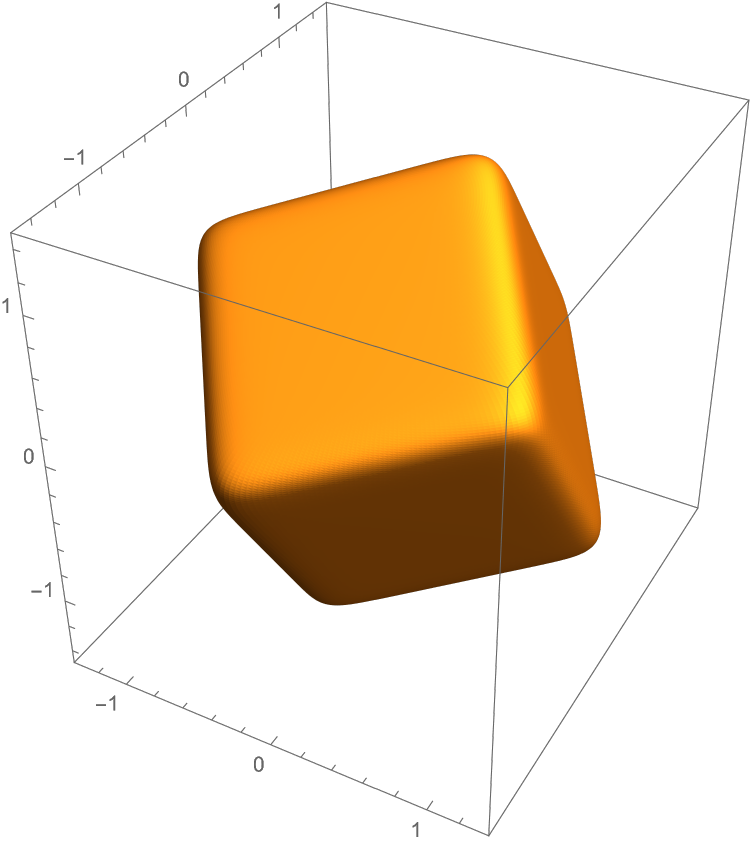}
}
\caption{Neural polyhedra of type $(3;p)$, with the activation function is chosen as $\left| x \right|^p$, where $p=0.6$, $0.8$, $0.9$, $1.0$, $1.2$, $1.5$, $2$, $3$, $5$, $10$ (from the left to the right).}
\label{fig6}
\end{center}
\end{figure*}

\begin{figure*}[!h]
\vskip 0.2in
\begin{center}
\centerline{
\includegraphics[width=0.2\columnwidth]{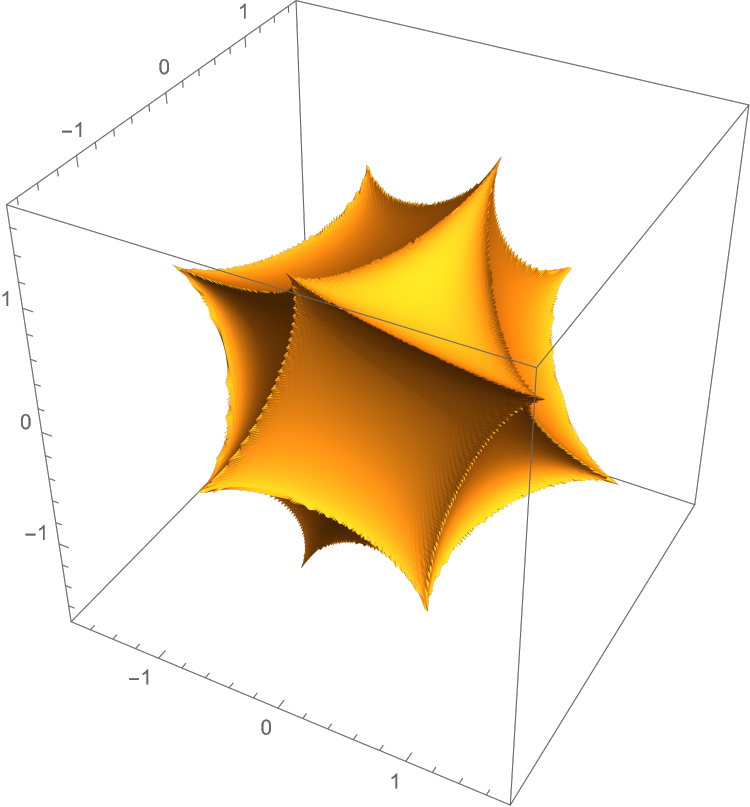}
\includegraphics[width=0.2\columnwidth]{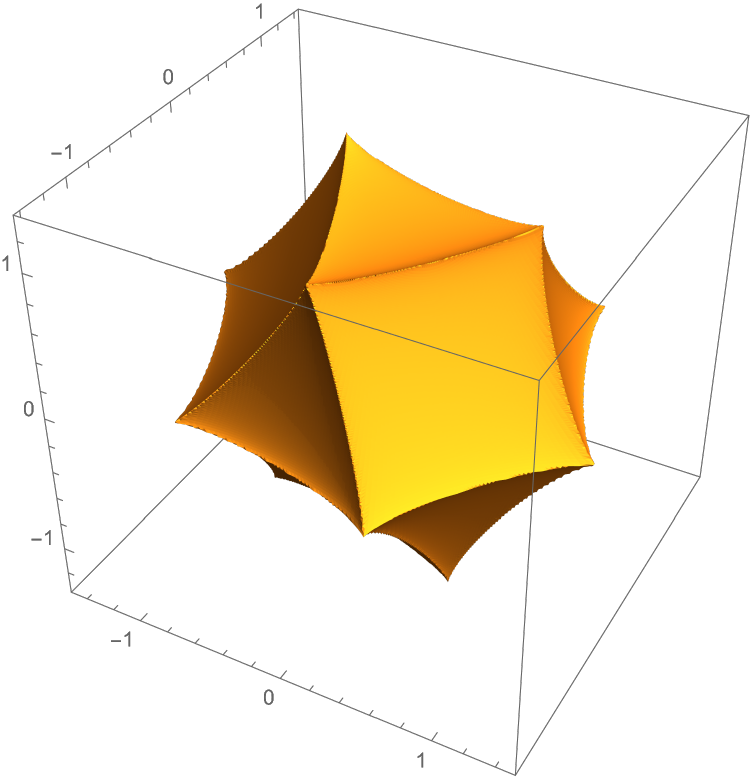}
\includegraphics[width=0.2\columnwidth]{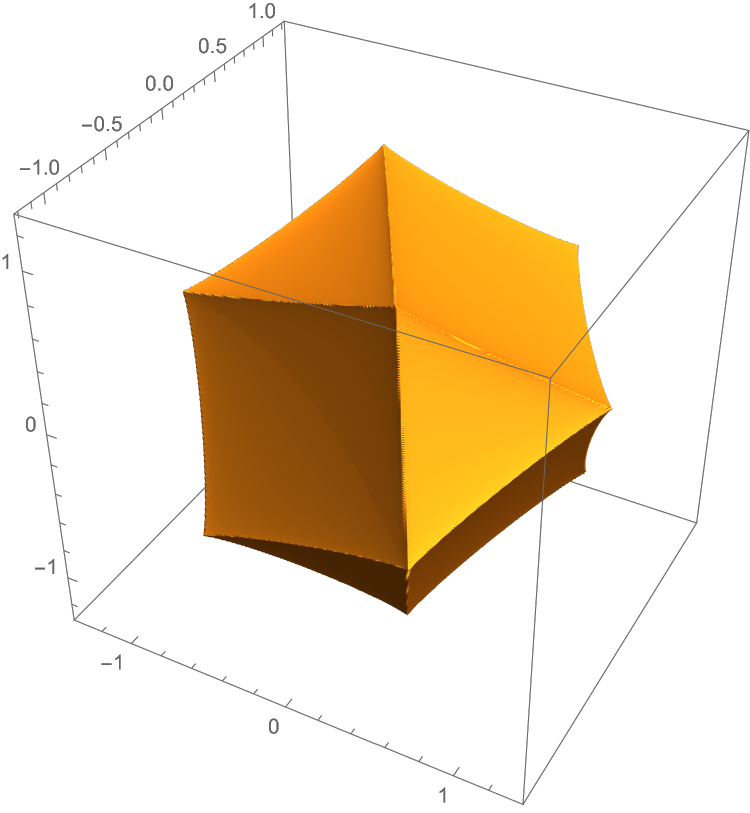}
\includegraphics[width=0.2\columnwidth]{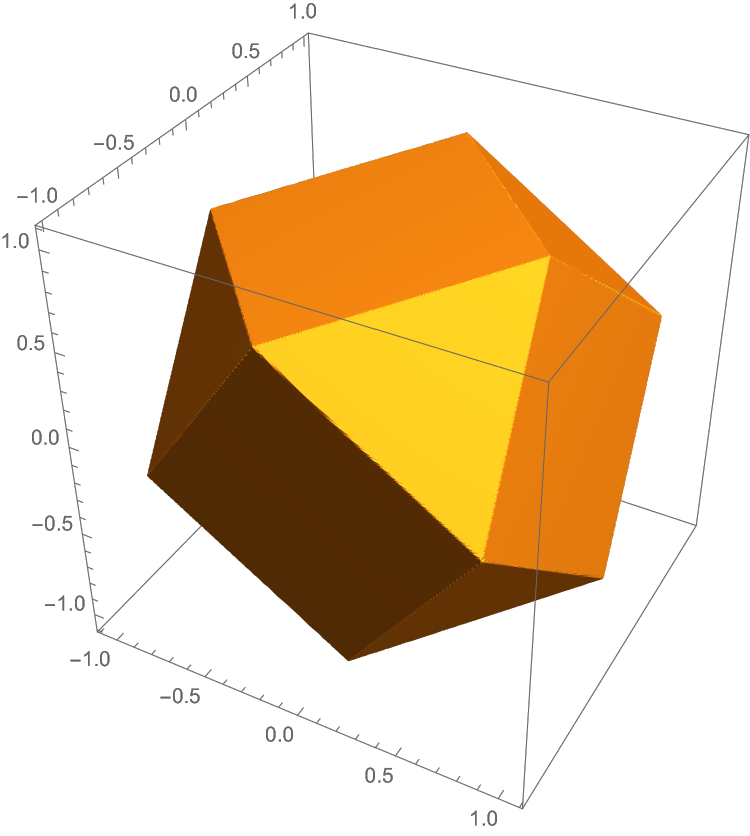}
\includegraphics[width=0.2\columnwidth]{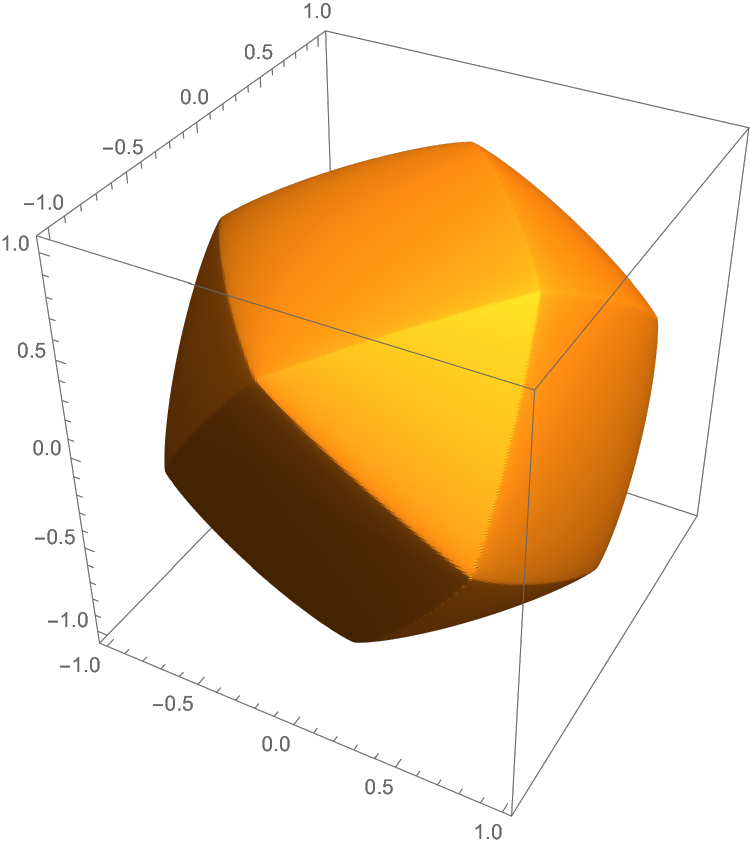}
\includegraphics[width=0.2\columnwidth]{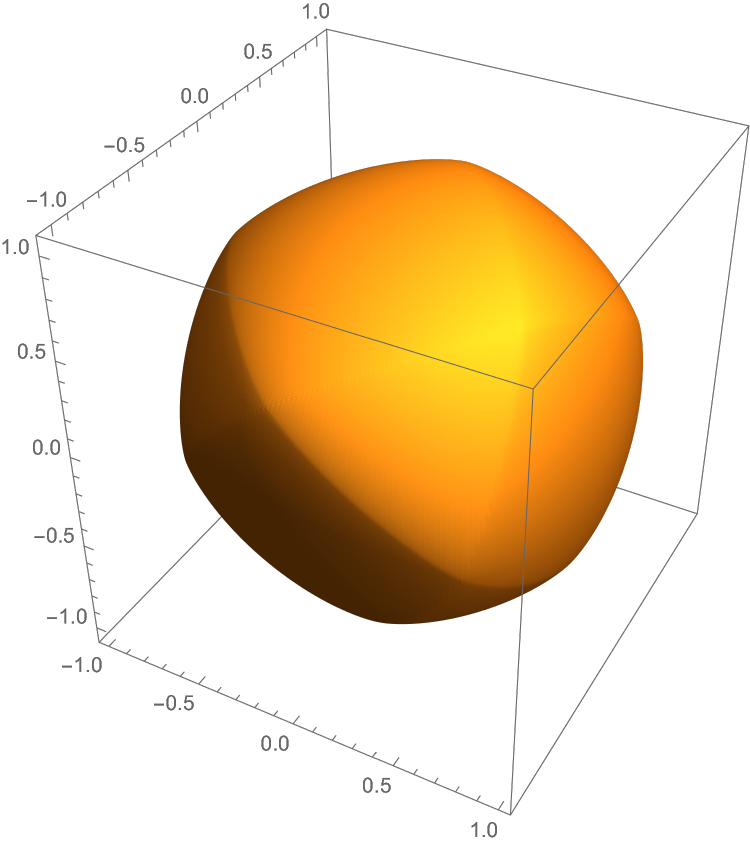}
\includegraphics[width=0.2\columnwidth]{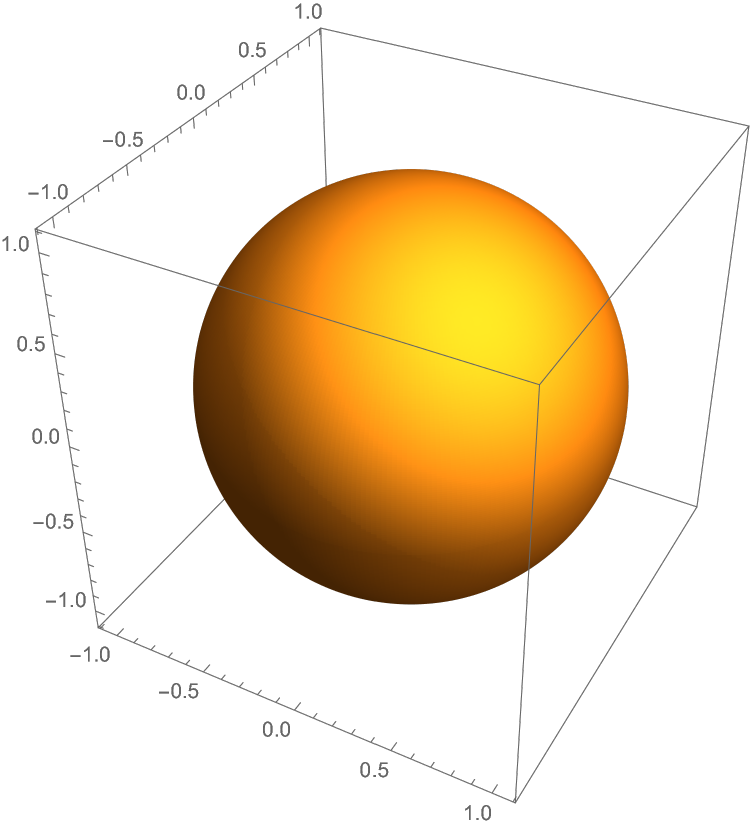}
\includegraphics[width=0.2\columnwidth]{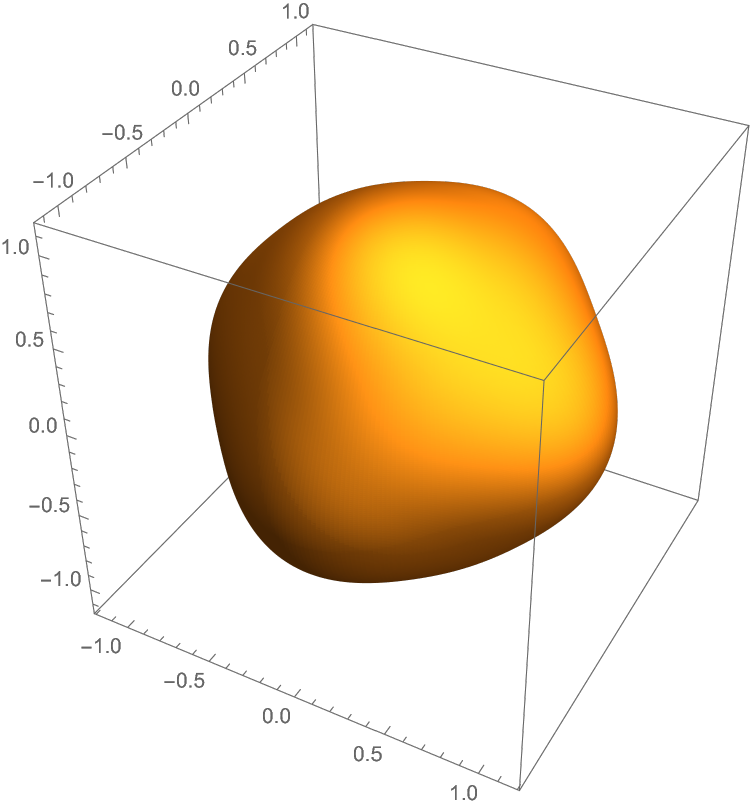}
\includegraphics[width=0.2\columnwidth]{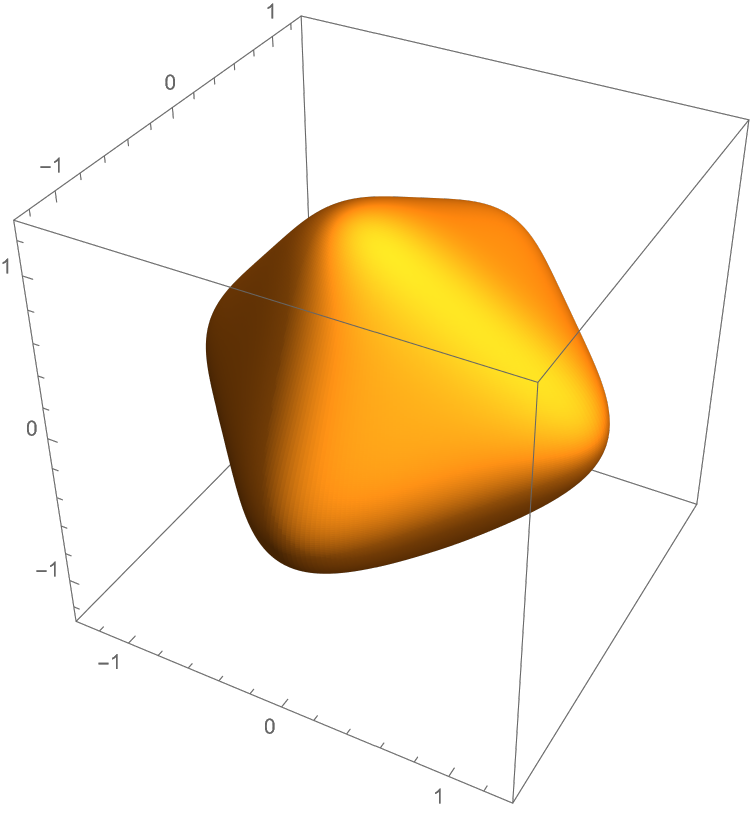}
\includegraphics[width=0.2\columnwidth]{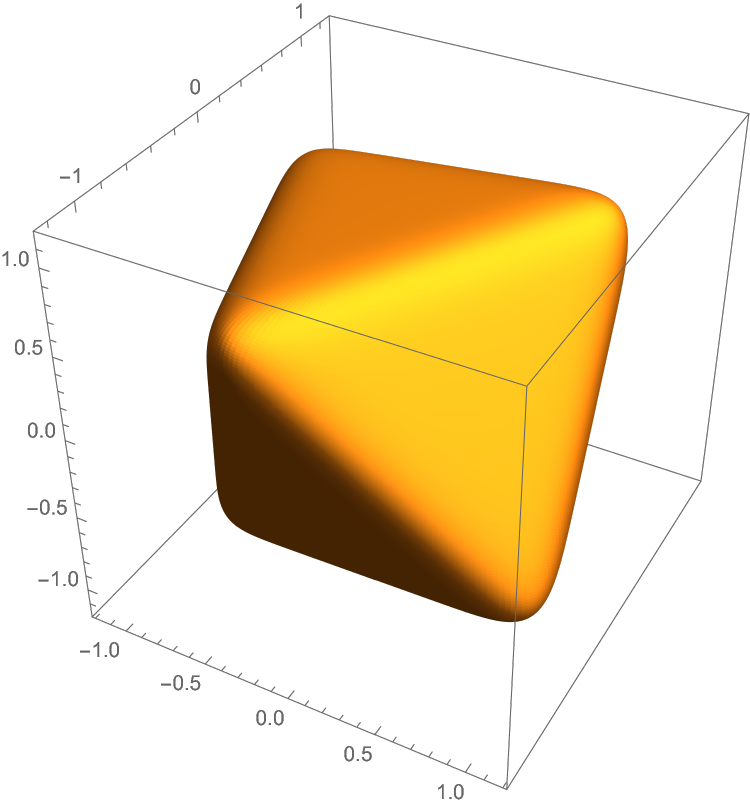}
}
\caption{Neural polyhedra of type $(4;p)$, with the activation function is chosen as $ \left| x \right|^p$, where $p=0.6$, $0.8$, $0.9$, $1.0$, $1.2$, $1.5$, $2$, $3$, $5$, $10$ (from the left to the right).}
\label{fig7}
\end{center}
\end{figure*}

\begin{figure*}[!h]
\begin{center}
\centerline{
\includegraphics[width=0.2\columnwidth]{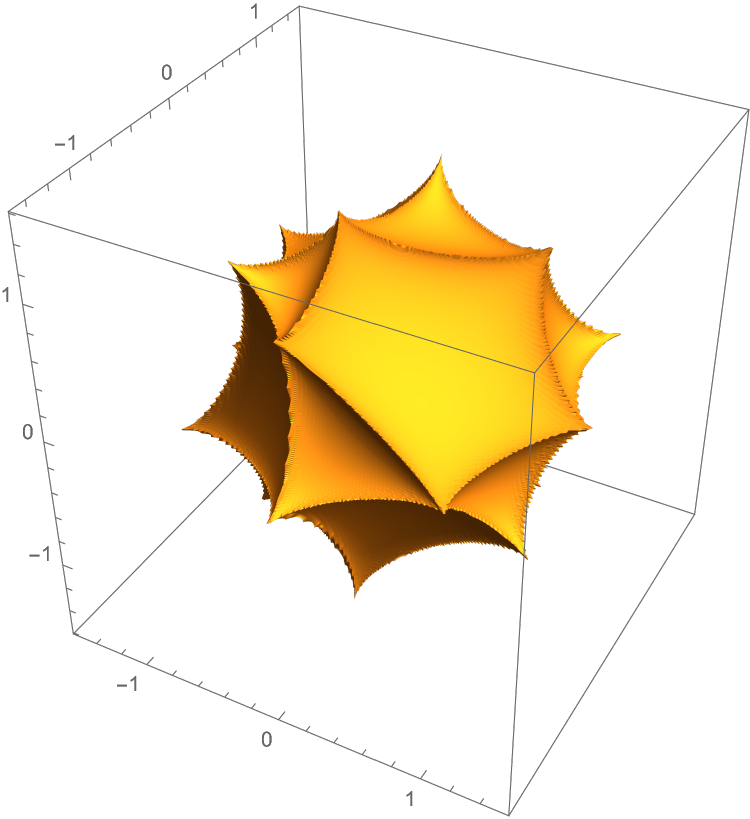}
\includegraphics[width=0.2\columnwidth]{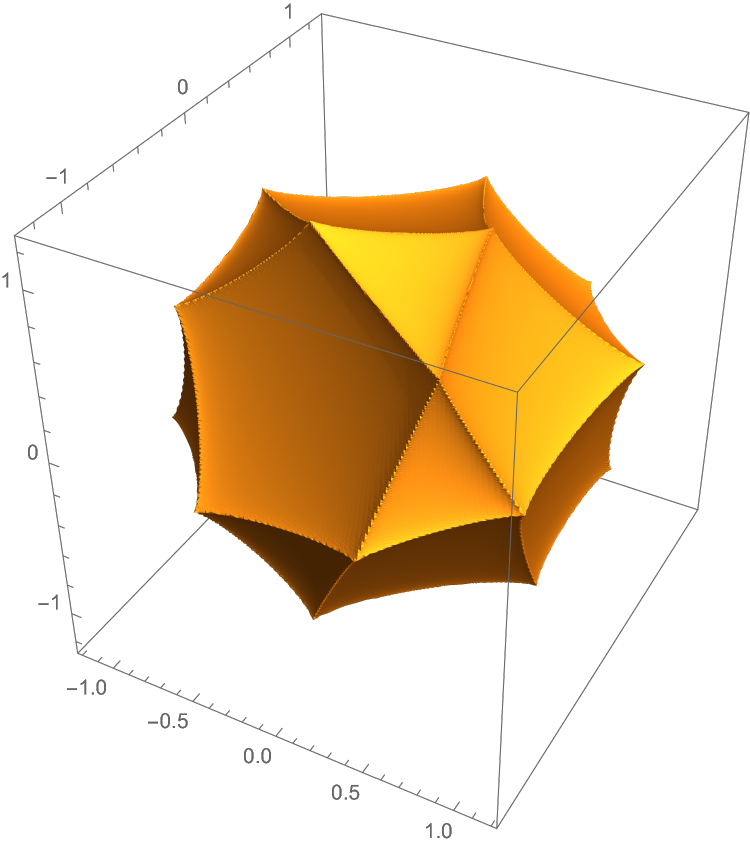}
\includegraphics[width=0.2\columnwidth]{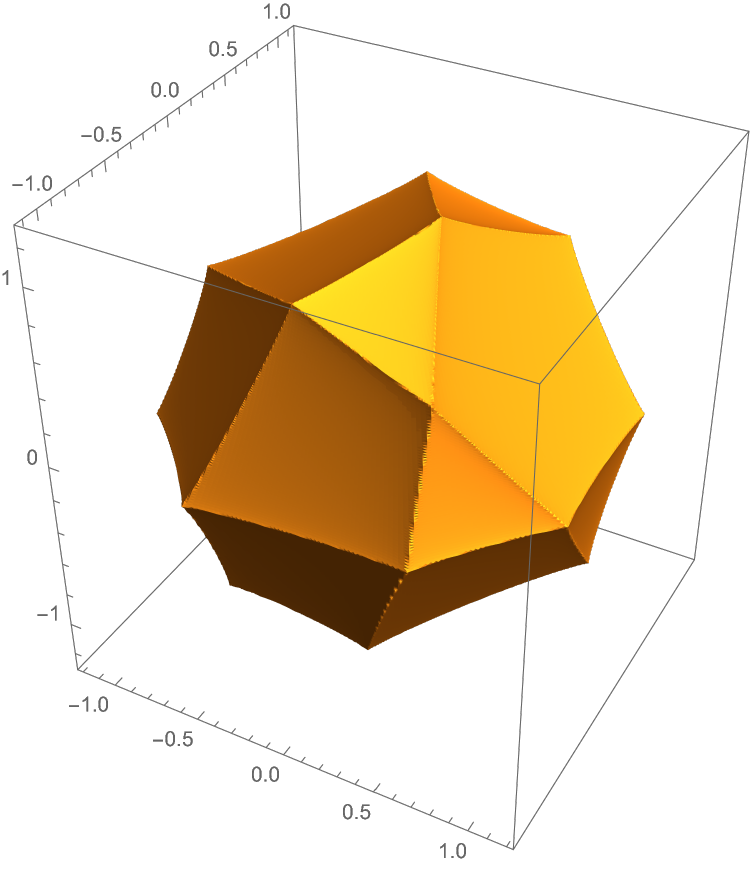}
\includegraphics[width=0.2\columnwidth]{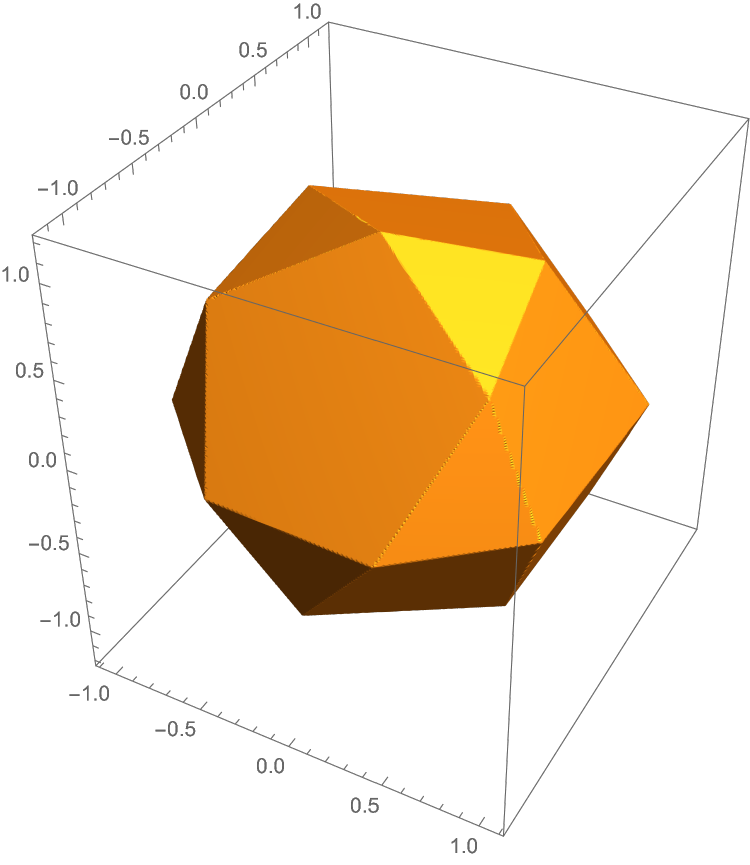}
\includegraphics[width=0.2\columnwidth]{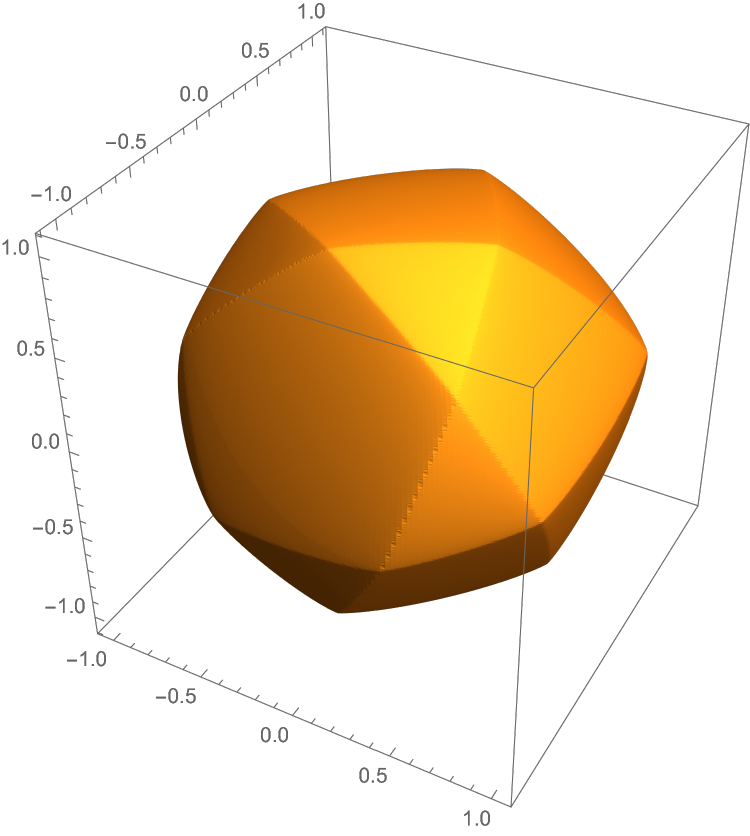}
\includegraphics[width=0.2\columnwidth]{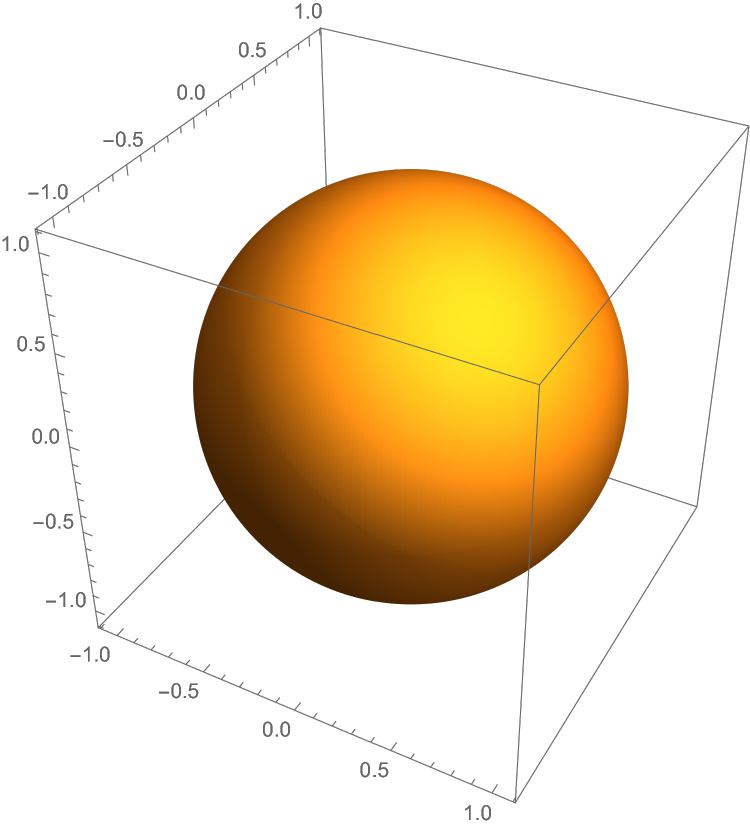}
\includegraphics[width=0.2\columnwidth]{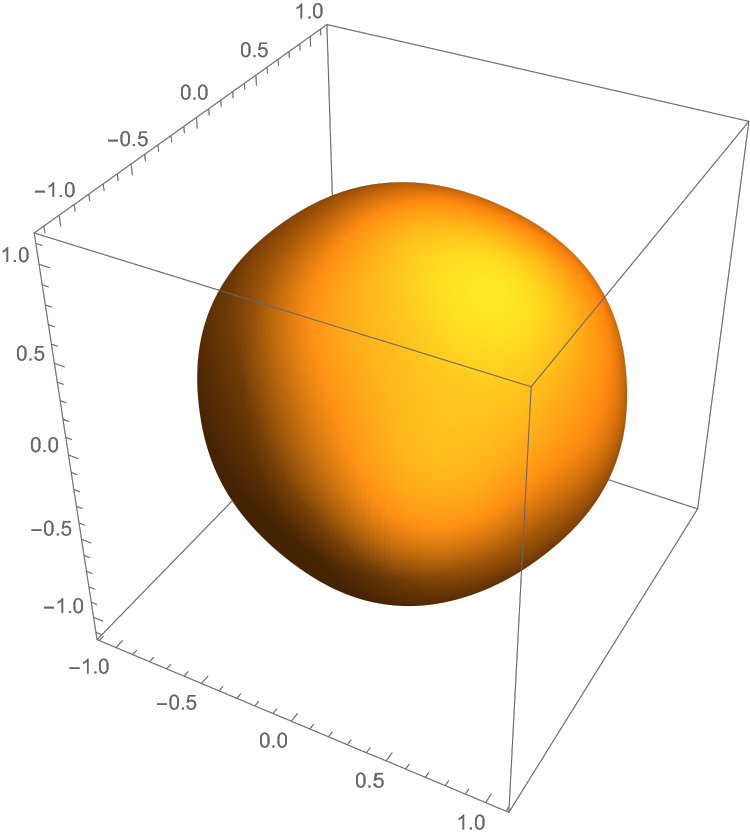}
\includegraphics[width=0.2\columnwidth]{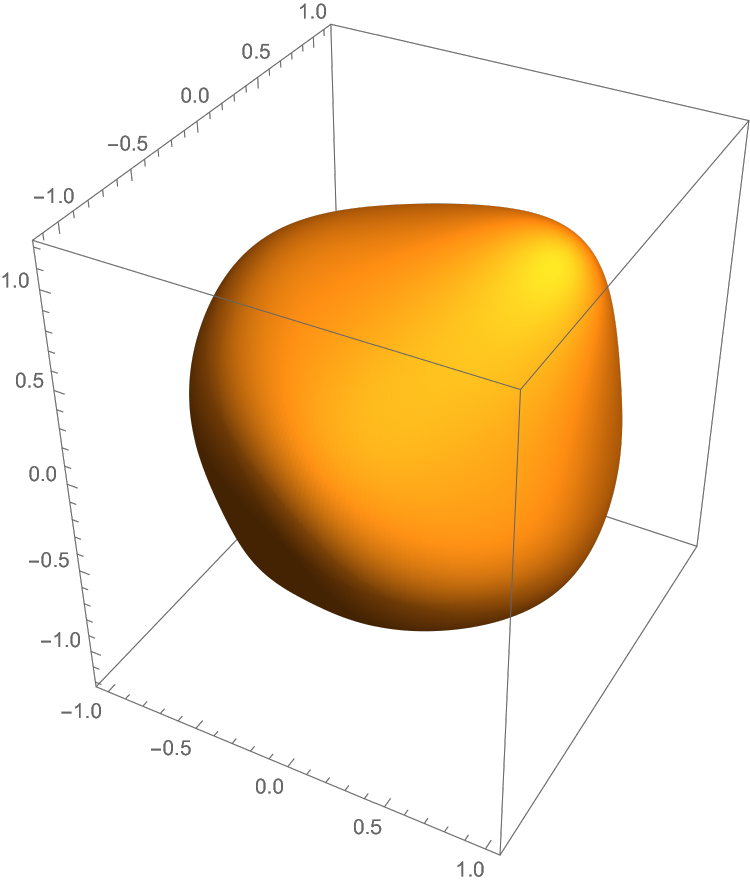}
\includegraphics[width=0.2\columnwidth]{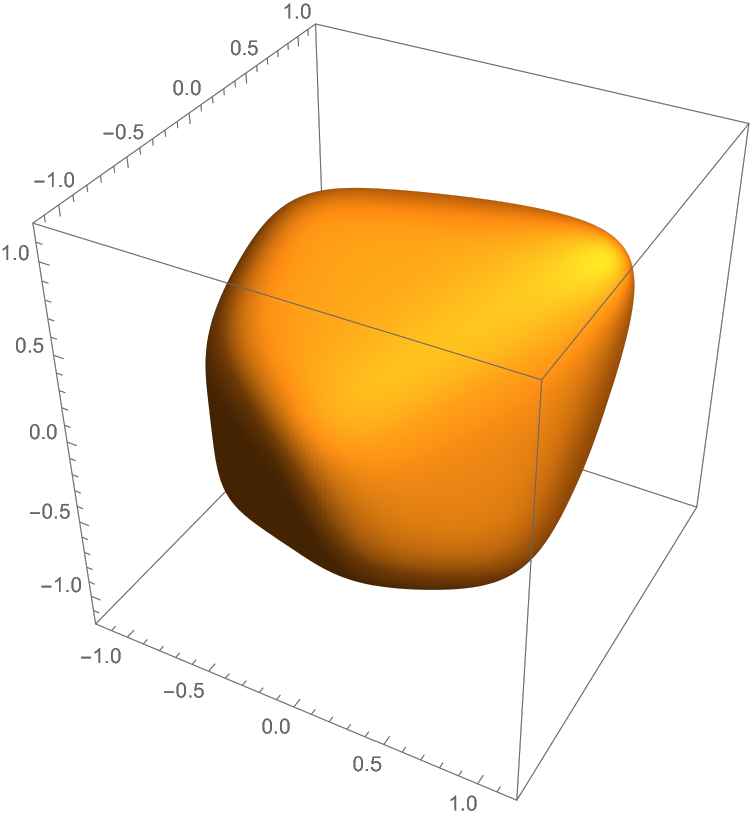}
\includegraphics[width=0.2\columnwidth]{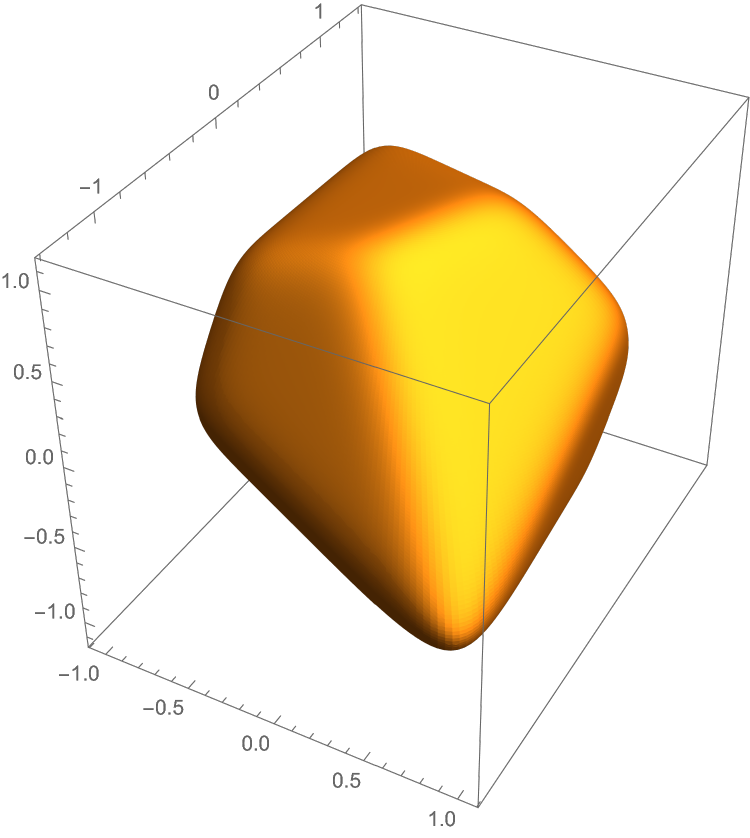}
}
\caption{Neural polyhedra of type $(5;p)$, with the activation function is chosen as $\left| x \right|^p$, where $p=0.6$, $0.8$, $0.9$, $1.0$, $1.2$, $1.5$, $2$, $3$, $5$, $10$ (from the left to the right).}
\label{fig8}
\end{center}
\vspace{0.3in}
\end{figure*}

\section*{Results}

In this report, we concentrate on the symmetric activation function \eqref{symact} to find rather symmetric neural polytopes.\footnote{Numerical training of the neural networks was done using Mathematica. The plots are numerical solutions of \eqref{eqfxi1} in spherical coordinates.} This also serves to test the ``rediscovery" of regular polygons and polyhedra for our choice $p=1$ in the activation function. 

\subsection*{Polygons, polyhedra and polytopes}

First, we report our results for $p=1$. In Fig.~\ref{fig1}, we show the results of the neural polygons. This is the $2$-polytopes of type $(n;1)$. It is amusing to find that $2n$-sided regular polygons known for thousands of years are reproduced beautifully, thus establishing the map between the neural network architecture and the regular polygons.

Next, we list the results of neural polyhedra. These are the neural $3$-polytopes of type $(n;1)$, see Fig.~\ref{fig2}. The type $(3;1)$ is the octahedron, which is one of the regular polyhedra. The type $(4;1)$ is the cuboctahedron, and the type $(6;1)$ is the icosidodecahedron, both of which are of the Archimedean solids. It is again amusing that various truncated polyhedra show up naturally.

Deeper neural network approximates the sphere more efficiently, and in Fig.~\ref{fig3} we list the results of neural $3$-polytope of type $(n,2;1)$. Compared with type $(n;1)$ with sharing the same $n$, we find that deeper network approximate the sphere better.

In Fig.~\ref{fig4}, we provide a higher-dimensional example: a neural $5$-polytope of type $(8;1)$. The plot is various 3-dimensional slices of the 5-polytope.

\subsection*{Neural polytopes}

Next, let us turn to the case $p\neq 1$. Interestingly, neural polytopes with $p\neq 1$ are spiky ($p<1$) or round ($1<p<\infty$) generalizations of the ordinary polytopes ($p=1$). In the limit $p \to\infty$, the neural polytopes become ordinary polytopes different from those for $p=1$.
At $p=2$, the neural polytopes are spheres.

First, in Fig.~\ref{fig5}, we show  neural 2-polytopes (neural polygons) of type $(2;1)$ with $p=0.8$, $1.0$, $1.2$, $1.5$, $2.0$, $3.0$, $5.0$ and $10.0$. At $p=1.0$ the neural polygon is a square.
Increasing $p$ makes the edge vertex rounded. At $p=2$ the neural polygon is a circle. Then, at $p=\infty$ the neural polygon becomes again a square.
The edge shape of the neural polygons is actually identical to the shape of the activation function $|x|^p$. 

Neural 3-polytopes (neural polyhedra) are plotted in Figs.~\ref{fig6}, \ref{fig7}, and \ref{fig8}. The behavior in change of the value of $p$ is quite similar to that of the neural polygons. 

For the neural polyhedra of type $(3;p)$, the $p=1$ neural polyhedron is an octahedron, while $p=\infty$ is a cube. Interestingly, these are dual polyhedra. So the $p=1$ and $p=\infty$ neural polyhedra provide a well-known duality among polytopes. This phenomenon is again natural in the sense that the activation function $\left| x \right|$ produces a kink at $x=0$, while $\left| x \right|^\infty$ is flat around $x=0$ and has kinky divergence at $x=\pm 1$.  
The neural polyhedra of type $(3;p)$ interpolates a polyhedron and a dual polyhedron, in between which a sphere appears. This duality also applies to the neural polygons of type $(2;p)$.\footnote{The neural polyhedra of type $(4;p)$ or $(5;p)$ shown in Fig.~\ref{fig7} or \ref{fig8} do not exhibit the precise duality, although close to it (the resultant discrete symmetries seem to be kept for generic values of $p$).}

\section*{Summary}

Polygons, polyhedra, and polytopes --- the fundamental objects in geometry --- were found to be a particular type of neural networks with linear activations and no bias. This amounts to visualization of an approximation of a sphere by neural network functions. We introduced neural polytopes, which are the natural consequence of generic activations. They round off edges of the ordinary polytopes, and include automatic interpolation of dual polytopes. The neural polytopes open the possibility of bridging discrete geometry and machine learning, and of even generalizing the discrete geometry, for geometric engineering of the nature.

\section*{Related work}

The geometric interpretation of ReLU networks as linearly segmented regions was studied in \cite{nair2010rectified}. 
The expressibility/complexity analysis of deep neural networks with piecewise linear activations is found in \cite{montufar2014number,pascanu2013number,arora2016understanding,balestriero2018mad,serra2018bounding,hanin2019complexity,croce2019provable,xu2022traversing,haase2023lower}, where the notion of partitioned regions are used.
The number of faces of our polytopes can be estimated in the large network approximation. Polyhedral theory was used \cite{arora2016understanding} for the complexity analysis.
The reverse engineering of ReLU networks was studied from the geometric viewpoint
in \cite{rolnick2020reverse}.
The polytope interpretation of the semanticity of neural networks was explored recently in 
\cite{black2022interpreting}. Our work focuses on the geometric and symmetric features of visualized neural network functions and the connection to discrete geometry.

Supervised learning of features of lattice polytopes was performed in \cite{bao2021polytopes, berglund2021machine, coates2023machine} and a genetic algorithm was used to generate lattice reflexive polytopes \cite{berglund2023new} for the analysis of Calabi-Yau manifolds.

\section*{Broader impact}

Polytopes are the fundamental objects in discrete geometry, whose applications
range from computer graphics to engineering and physics. Our finding bridges 
the discrete geometry with neural networks directly, with which we expect  automatic generation of discrete geometries from point data of natural surfaces,
together with a deeper mathematical understanding of neural network functions.
This work unites philosophically the distinct two ideas of the human visual recognition --- the polytopes to model the nature by the ancient Greek mathematics of Archimedes, and the neural network functions to model any shape in the nature. This occurrence could be called a societal Poincar\'e recurrence.


\section*{Acknowledgements}
K.~H.~is indebted to Koji Miyazaki for valuable discussions, education and support.
The work of K.~H.~was supported in part by JSPS KAKENHI Grant Nos.~JP22H01217, JP22H05111, and JP22H05115.
The work of T.~N.~was supported in part by JSPS KAKENHI Grant Nos.~JP22K20372, JP23H04526, JP23H01845, and JP23K03426.
The work of H.~N.~was supported in part by JSPS KAKENHI Grant Nos.~JP19K03488 and JP23H01072.

\bibliography{paper}
\bibliographystyle{synsml2023}



\end{document}